\documentclass[10pt,journal,compsoc]{IEEEtran}

 \usepackage{amsmath}
\usepackage{amssymb}
\usepackage{graphicx}
\usepackage{array}
\usepackage{float}
\usepackage{stfloats}
\usepackage{bm}
\usepackage{booktabs}
\usepackage{colortbl}
\usepackage{times}
\usepackage{epsfig}
\usepackage[latin9]{inputenc}
\usepackage{mathrsfs}
\usepackage{hyperref}

\providecommand{\tabularnewline}{\\}
\floatstyle{ruled}
\newfloat{algorithm}{tbp}{loa}
\providecommand{\algorithmname}{Algorithm}
\floatname{algorithm}{\protect\algorithmname}

\ifCLASSOPTIONcompsoc
  \usepackage[nocompress]{cite}
\else
  \usepackage{cite}
\fi
\ifCLASSINFOpdf
\else
\fi
\begin{document}
%
\title{shapeDTW: shape Dynamic Time Warping}
%
%
%
%

\author{Jiaping~Zhao,~\IEEEmembership{student Member,~IEEE}
        Laurent~Itti,~\IEEEmembership{Member,~IEEE}    
\IEEEcompsocitemizethanks{\IEEEcompsocthanksitem J. Zhao is with the Department
of Computer Science, University of Southern California, Los Angeles,
CA, 90089.\protect\\
E-mail: jiapingz@usc.edu
\IEEEcompsocthanksitem L. Itti is with the Department
of Computer Science, University of Southern California., Los Angeles,
CA, 90089.\protect\\
E-mail: itti@usc.edu}
\thanks{Manuscript, June, 2016.}}

%
%

\markboth{manuscript TKDE}%
{Shell \MakeLowercase{\textit{et al.}}: Bare Advanced Demo of IEEEtran.cls for Journals}
%



\IEEEtitleabstractindextext{%
\begin{abstract}
Dynamic Time Warping (DTW) is an algorithm to align temporal sequences
with possible local non-linear distortions, and has been widely applied
to audio, video and graphics data alignments. DTW is essentially a
point-to-point matching method under some boundary and temporal consistency
constraints. Although DTW obtains a global optimal solution, it does
not necessarily achieve locally sensible matchings. Concretely, two
temporal points with entirely dissimilar local structures may be matched
by DTW. To address this problem, we propose an improved alignment
algorithm, named shape Dynamic Time Warping (shapeDTW), which enhances
DTW by taking point-wise local structural information into consideration.
shapeDTW is inherently a DTW algorithm, but additionally attempts
to pair locally similar structures and to avoid matching points with
distinct neighborhood structures. We apply shapeDTW to align audio
signal pairs having ground-truth alignments, as well as artificially
simulated pairs of aligned sequences, and obtain quantitatively much
lower alignment errors than DTW and its two variants. When shapeDTW
is used as a distance measure in a nearest neighbor classifier (NN-shapeDTW)
to classify time series, it beats DTW on 64 out of 84 UCR time series
datasets, with significantly improved classification accuracies. By
using a properly designed local structure descriptor, shapeDTW improves
accuracies by more than $10\%$ on 18 datasets. To the best of our
knowledge, shapeDTW is the first distance measure under the nearest
neighbor classifier scheme to significantly outperform DTW, which
had been widely recognized as the best distance measure to date. Our code is publicly accessible at: \url{https://github.com/jiapingz/shapeDTW}.
\end{abstract}

}

\maketitle

\IEEEdisplaynontitleabstractindextext

%
\IEEEpeerreviewmaketitle

\ifCLASSOPTIONcompsoc
\IEEEraisesectionheading{\section{Introduction}\label{sec:introduction}}
\else
\section{Introduction}
\label{sec:introduction}
\fi

%
%
%
%

\begin{figure}[htbp!]
\begin{centering}
\includegraphics[width=0.42\textwidth]{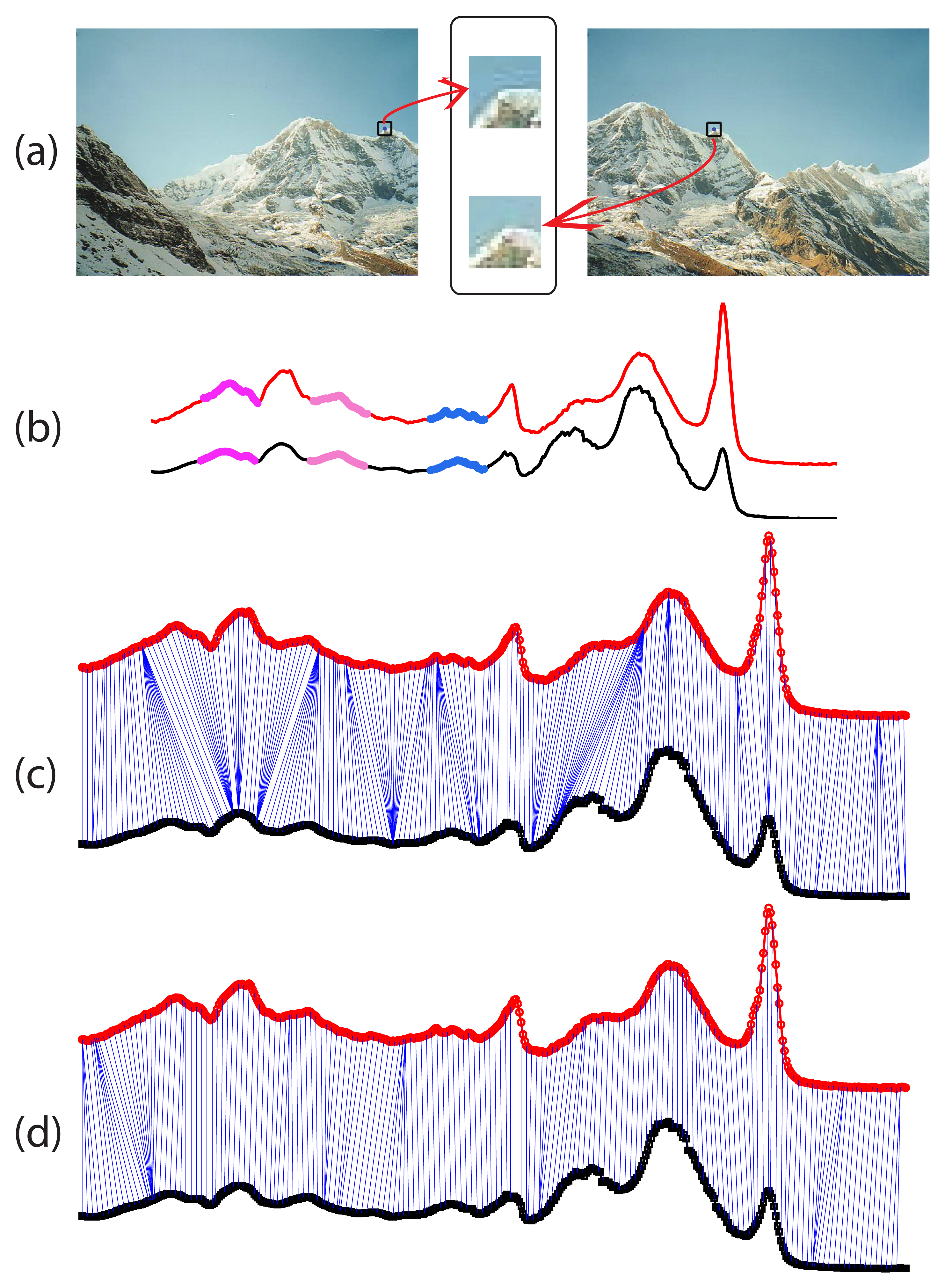} 
\par\end{centering}

\caption{\label{fig:motivation} Motivation to incorporate temporal neighborhood
structural information into the sequence alignment process. (a) an
image matching example: two corresponding points from the image pairs
are boxed out and their local patches are shown in the middle. Local
patches encode image structures around spatial neighborhoods, and
therefore are discriminative for points, while it is hard to match
two points solely by their pixel values. (b) two time series with
several similar local structures, highlighted as bold segments. (c)
DTW alignment: DTW fails to align similar local structures. (d) shapeDTW
alignment: we achieve a more interpretable alignment, with similarly-shaped
local structures matched.}
\end{figure}

\begin{figure*}[htbp!]
\begin{centering}
\includegraphics[width=0.8\textwidth]{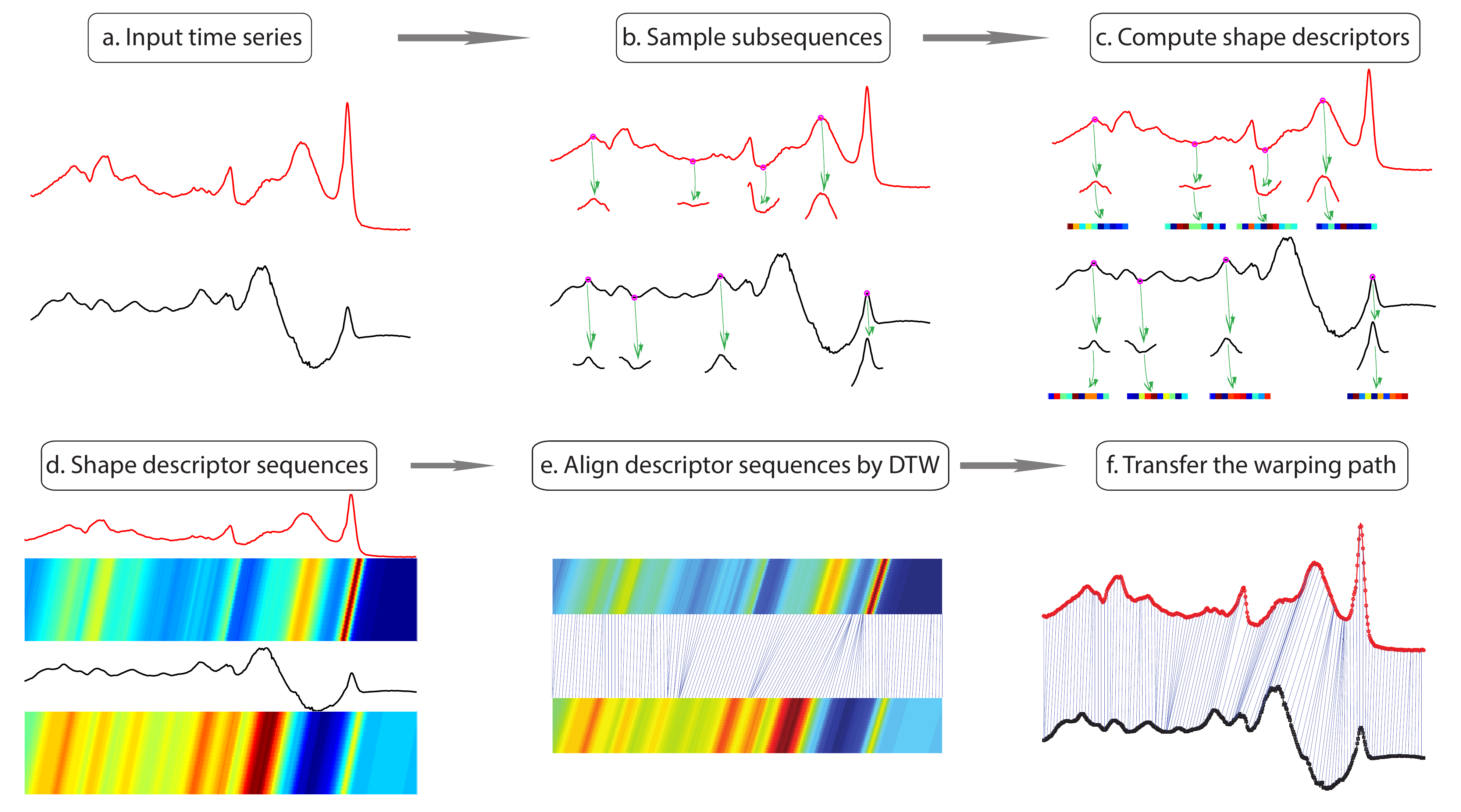} 
\par\end{centering}

\caption{\label{fig:flowchart} Pipeline of shapeDTW. shapeDTW consists of
two major steps: encode local structures by shape descriptors and
align descriptor sequences by DTW. Concretely, we sample a subsequence
from each temporal point, and further encode it by some shape descriptor.
As a result, the original time series is converted into a descriptor
sequence of the same length. Then we align two descriptor sequences
by DTW and transfer the found warping path to the original time series. }
\end{figure*}

\IEEEPARstart{D}{ynamic} time warping (DTW) is an algorithm to
align temporal sequences, which has been widely used in speech recognition
\cite{rabiner1993fundamentals}, human motion animation \cite{hsu2005style},
human activity recognition \cite{kulkarni2014continuous} and time
series classification \cite{UCRArchive}. DTW allows temporal sequences
to be locally shifted, contracted and stretched, and under some boundary
and monotonicity constraints, it searches for a global optimal alignment
path. DTW is essentially a point-to-point matching algorithm, but
it additionally enforces temporal consistencies among matched point
pairs. If we distill the matching component from DTW, the matching
is executed by checking the similarity of two points based on their
Euclidean distance. Yet, matching points based solely on their coordinate
values is unreliable and prone to error, therefore, DTW may generate
perceptually nonsensible alignments, which wrongly pair points with
distinct local structures (see Fig.\ref{fig:motivation} (c)). This
partially explains why the nearest neighbor classifier under the DTW
distance measure is less interpretable than the shapelet classifier
\cite{ye2009time}: although DTW does achieve a global minimal score,
the alignment process itself takes no local structural information
into account, possibly resulting in an alignment with little semantic
meaning. In this paper, we propose a novel alignment algorithm, named
shape Dynamic Time Warping (shapeDTW), which enhances DTW by incorporating
point-wise local structures into the matching process. As a result,
we obtain perceptually interpretable alignments: similarly-shaped
structures are preferentially matched based on their degree of similarity.
We further quantitatively evaluate alignment paths against the ground-truth
alignments, and shapeDTW achieves much lower alignment errors than
DTW on both simulated and real sequence pairs. An alignment example
by shapeDTW is shown in Fig.\ref{fig:motivation} (d).

Point matching is a well studied problem in the computer vision community,
widely known as image matching. In order to search corresponding points
from two distinct images taken from the same scene, a quite naive
way is to compare their pixel values. But pixel values at a point
lacks spatial neighborhood context, making it less discriminative
for that point; e.g., a tree leaf pixel from one image may have exactly
the same RGB values as a grass pixel from the other image, but these
two pixels are not corresponding pixels and should not be matched.
Therefore, a routine for image matching is to describe points by their
surrounding image patches, and then compare the similarities of point
descriptors. Since point descriptors designed in this way encode image
structures around local neighborhoods, they are more distinctive and
discriminative than single pixel values. In early days, raw image
patches were used as point descriptors \cite{agarwal2002learning},
and now more powerful descriptors like SIFT \cite{lowe2004distinctive}
are widely adopted since they capture local image structures very
well and are invariant to image scale and rotation.

Intuitively, local neighborhood patches make points more discriminative
from other points, while matching based on RGB pixel values is brittle
and results in high false positives. However, the matching component
in the traditional DTW bears the same weakness as image matching based
on single pixel values, since similarities between temporal points
are measured by their coordinates, instead of by their local neighborhoods.
An analogous remedy for temporal matching hence is: first encode each
temporal point by some descriptor, which captures local subsequence
structural information around that point, and then match temporal
points based on the similarity of their descriptors. If we further
enforce temporal consistencies among matchings, then comes the algorithm
proposed in the paper: shapeDTW.

shapeDTW is a temporal alignment algorithm, which consists of two
sequential steps: (1) represent each temporal point by some shape
descriptor, which encodes structural information of local subsequences
around that point; in this way, the original time series is converted
into a sequence of descriptors. (2) use DTW to align two sequences
of descriptors. Since the first step takes linear time while the second
step is a typical DTW, which takes quadratic time, the total time
complexity is quadratic, indicating that shapeDTW has the same computational
complexity as DTW. However, compared with DTW and its variants (derivative
Dynamic Time Warping (dDTW) \cite{keogh2001derivative} and weighted
Dynamic Time Warping (wDTW)\cite{jeong2011weighted}), it has two
clear advantages: (1) shapeDTW obtains lower alignment errors than
DTW/dDTW/wDTW on both artificially simulated aligned sequence pairs
and real audio signals; (2) the nearest neighbor classifier under
the shapeDTW distance measure (NN-shapeDTW) significantly beats NN-DTW
on 64 out of 84 UCR time series datasets \cite{UCRArchive}. NN-shapeDTW
outperforms NN-dDTW/NN-wDTW significantly as well. Our shapeDTW time series 
alignment procedure is shown in Fig. \ref{fig:flowchart}.

Extensive empirical experiments have shown that a nearest neighbor
classifier with the DTW distance measure (NN-DTW) is the best choice
to date for most time series classification problems, since no alternative
distance measures outperforms DTW significantly \cite{wang2013experimental,rakthanmanon2012searching,petitjean2014dynamic}.
However, in this paper, the proposed temporal alignment algorithm,
shapeDTW, if used as a distance measure under the nearest neighbor
classifier scheme, significantly beats DTW. To the best of our knowledge,
shapeDTW is the first distance measure that outperforms DTW significantly.

Our contributions are several fold: (1) we propose a temporal alignment
algorithm, shapeDTW, which is as efficient as DTW (dDTW, wDTW) but
achieves quantitatively better alignments than DTW (dDTW, wDTW); (2)
Working under the nearest neighbor classifier as a distance measure
to classify 84 UCR time series datasets, shapeDTW, under all tested
shape descriptors, outperforms DTW significantly; (3) shapeDTW provides
a quite generic alignment framework, and users can design new shape
descriptors adapted to their domain data characteristics and then
feed them into shapeDTW for alignments.

\section{Related work}

Since shapeDTW is developed for sequence alignment, here we first
review research work related to sequence alignment. DTW is a typical
sequence alignment algorithm, and there are many ways to improve DTW
to obtain better alignments. Traditionally, we could enforce global
warping path constraints to prevent pathological warpings \cite{rabiner1993fundamentals},
and several typical such global warping constraints include Sakoe-Chiba
band and Itakura Parallelogram. Similarly, we could choose to use
different step patterns in different applications: apart from the
widely used step pattern - ``symmetric1'', there are other popular steps
patterns like ``symmetric2'', ``asymmetric'' and ``RabinerJuangStepPattern''
\cite{giorgino2009computing}. However, how to choose an appropriate
warping band constraint and a suitable step pattern depends on our
prior knowledge on the application domains.

There are several recent works to improve DTW alignment. In \cite{keogh2001derivative},
to get the intuitively correct ``feature to feature'' alignment
between two sequences, the authors introduced derivative dynamic time
warping (dDTW), which computes first-order derivatives of time series
sequences, and then aligns two derivative sequences by DTW. In \cite{jeong2011weighted},
the authors developed weighted DTW (wDTW), which is a penalty-based
DTW. wDTW takes the phase difference between two points into account
when computing their distances. Batista et al \cite{batista2011complexity}
proposed a complexity-invariant distance measure, which essentially
rectifies an existing distance measure (e.g., Euclidean, DTW) by multiplying
a complexity correction factor. Although they achieve improved results
on some datasets by rectifying the DTW measure, they do not modify
the original DTW algorithm. In \cite{lajugie2014metric}, the authors
proposed to learn a distance metric, and then align temporal sequences
by DTW under this new metric. One major drawback is the requirement
of ground truth alignments for metric learning, because in reality
true alignments are usually unavailable. In \cite{candan2012sdtw},
the authors proposed to utilize time series local structure information
to constrain the search of the warping path. They introduce a SIFT-like
feature point detector and descriptor to detect and match salient
feature points from two sequences first, and then use matched point
pairs to regularize the search scope of the warping path. Their major
initiative is to improve the computational efficiency of dynamic time
warping by enforcing band constraints on the potential warping paths,
such that they do not have to compute the full accumulative distance
matrix between the two sequences. Our method is sufficiently different
from theirs in following aspects: first, we have no notion of feature
points, while feature points are key to their algorithm, since feature
points help to regularize downstream DTW; second, our algorithm aims
to achieve better alignments, while their algorithm attempts to improve
the computational efficiency of the traditional DTW. In \cite{petitjean2014dynamic},
the authors focus on improving the efficiency of the nearest neighbor
classifier under the DTW distance measure, but they keep the traditional
DTW algorithm unchanged.

Our algorithm, shapeDTW, is different from the above works in that:
we measure similarities between two points by computing similarities
between their local neighborhoods, while all the above works compute
the distance between two points based on their single-point y-values
(derivatives).

Since shapeDTW can be applied to classify time series (e.g., NN-shapeDTW),
we review representative time series classification algorithms. In
\cite{lin2012rotation}, the authors use the popular Bag-of-Words
to represent time series instances, and then classify the representations
under the nearest neighbor classifier. Concretely, it discretizes
time series into local SAX \cite{lin2003symbolic} words, and uses
the histogram of SAX words as the time series representation. In \cite{rakthanmanon2013fast},
the authors developed an algorithm to first extract class-membership
discriminative shapelets, and then learn a decision tree classifier
based on distances between shapelets and time series instances. In
\cite{silva2013time}, they first represent time series using recurrent
plots, and then measure the similarity between recurrence plots using
Campana-Keogh (CK-1) distance (PRCD). PRCD distance is used as the
distance measure under the one-nearest neighbor classifier to do classification.
In \cite{baydogan2013bag}, a bag-of-feature framework to classify
time series is introduced. It uses a supervised codebook to encode
time series instances, and then uses random forest classifier to classify
the encoded time series. In \cite{grabocka2014invariant}, the authors
first encode time series as a bag-of-patterns, and then use polynomial
kernel SVM to do the classification. Zhao and Itti \cite{zhao2015classifying} proposed 
to first encode time series by the 2nd order encoding method - Fisher Vectors, and then classify 
encoded time series by a linear kernel SVM. In their paper, subsequences are sampled from both feature 
points and flat regions.

shapeDTW is different from above works in that: shapeDTW is developed
to align temporal sequences, but can be further applied to classify
time series. However, all above works are developed to classify time
series, and they are incapable to align temporal sequences at their
current stages. Since time series classification is only one application
of shapeDTW, we compare NN-shapeDTW against the above time series
classification algorithms in the supplementary materials.

The paper is organized as follows: the detailed algorithm for shapeDTW
is introduced in Sec.\ref{sec:shape-DTW}, and in Sec.\ref{sec:shape-descriptors}
we introduce several local shape descriptors. Then we extensively
test shapeDTW for both sequence alignments and time series classification
in Sec. \ref{sec:Experimental-validation}, and conclusions are drawn
in Sec.\ref{sec:Conclusion}.

\section{\label{sec:shape-DTW}shape Dynamic Time Warping}

In this section, we introduce a temporal alignment algorithm, shapeDTW.
First we introduce DTW briefly.

\subsection{Dynamic Time Warping}

DTW is an algorithm to search for an optimal alignment between two
temporal sequences. It returns a distance measure for gauging similarities
between them. Sequences are allowed to have local non-linear distortions
in the time dimension, and DTW handles local warpings to some extent.
DTW is applicable to both univariate and multivariate time series,
and here for simplicity we introduce DTW in the case of univariate
time series alignment.

A univariate time series $\mathcal{T}$ is a sequence of real values,
i.e., $\mathcal{T}=(t_{1},\, t_{2},...,\, t_{L})^{T}$.
Given two sequences $\mathcal{P}$ and $\mathcal{Q}$ of possible
different lengths $\mathcal{L_{P}}$ and $\mathcal{L_{Q}}$, namely
$\mathcal{P}=(p_{1},p_{2},...,p_{\mathcal{L_{P}}})^{T}$ and $\mathcal{Q}=(q_{1},q_{2},...,q_{\mathcal{L_{Q}}})^{T}$,
and let $\mathcal{D}(\mathcal{P},\mathcal{Q})\in\mathcal{R^{L_{P}\times L_{Q}}}$
be an pairwise distance matrix between sequences $\mathcal{P}$ and
$\mathcal{Q}$, where $\mathcal{D}(\mathcal{P},\mathcal{Q})_{i,j}$
is the distance between $p_{i}$ and $p_{j}$. One widely used pairwise
distance measure is the Euclidean distance, i.e., $\mathcal{D}(\mathcal{P},\mathcal{Q})_{i,j}=|p_{i}-q_{j}|$.
The goal of temporal alignment between $\mathcal{P}$ and $\mathcal{Q}$
is to find two sequences of indices $\alpha$ and $\beta$ of the
same length $l\,(l\geq\max(\mathcal{L_{P}},\mathcal{L_{Q}}))$, which
match index $\alpha(i)$ in the time series $\mathcal{P}$ to index
$\beta(i)$ in the time series $\mathcal{Q}$, such that the total
cost along the matching path $\sum_{i=1}^{l}\mathcal{D}(\mathcal{P},\mathcal{Q})_{\alpha(i),\beta(i)}$
is minimized. The alignment path $(\alpha,\beta)$ is constrained
to satisfies boundary, monotonicity and continuity conditions \cite{sakoe1978dynamic,keogh2005exact,garreau2014metric}:

\vspace{-12pt}

\begin{equation}
\begin{cases}
\begin{array}{l}
\alpha(1)=\beta(1)=1\\
\alpha(l)=\mathcal{L_{P}},\:\beta(l)=\mathcal{L_{Q}}\\
\left(\alpha(i+1),\beta(i+1))-(\alpha(i),\beta(i)\right)\in\left\{ (1,0),(1,1),(0,1)\right\} 
\end{array}\end{cases}\label{eq:DTWconditions}
\end{equation}

Given an alignment path $(\alpha,\beta)$, we define two warping matrices
$\mathcal{W^{P}}\in{\{0,1\}}^{l\times\mathcal{L_{P}}}$ and $\mathcal{W^{Q}}\in\{0,1\}^{l\times\mathcal{L_{Q}}}$
for $\mathcal{P}$ and $\mathcal{Q}$ respectively, such that $\mathcal{W^{P}}(i,\alpha(i))=1$,
otherwise $\mathcal{W^{P}}(i,j)=0$, and similarly $\mathcal{W^{Q}}(i,\beta(i))=1$,
otherwise $\mathcal{W^{Q}}(i,j)=0$. Then the total cost along the
matching path $\sum_{i=1}^{l}\mathcal{D}(\mathcal{P},\mathcal{Q})_{\alpha(i),\beta(i)}$
is equal to $\parallel\mathcal{W^{P}}\cdot\mathcal{P}-\mathcal{W^{Q}}\cdot\mathcal{Q}\parallel_{1}$,
thus searching for the optimal temporal matching can be formulated
as the following optimization problem:

\vspace{-12pt}

\begin{equation}
{\arg\min}_{l,\,\mathcal{W^{P}}\in\{0,1\}^{l\times\mathcal{L_{P}}},\,\mathcal{W^{Q}}\in\{0,1\}^{l\times\mathcal{L_{Q}}}}\parallel\mathcal{W^{P}}\cdot\mathcal{P}-\mathcal{W^{Q}}\cdot\mathcal{Q}\parallel_{1}\label{eq:DTW}
\end{equation}

Program \ref{eq:DTW} can be solved efficiently in $\mathcal{O}(\mbox{\ensuremath{\mathcal{L_{P}}\times}\ensuremath{\ensuremath{\mathcal{L_{Q}}}}})$
time by a dynamic programming algorithm \cite{EllisDTW}. Various
different moving patterns and temporal window constraints \cite{sakoe1978dynamic}
can be enforced, but here we consider DTW without warping window constraints
and taking moving patterns as in (\ref{eq:DTWconditions}).

\subsection{shape Dynamic Time Warping}

DTW finds a global optimal alignment under certain constraints, but
it does not necessarily achieve locally sensible matchings. Here we
incorporate local shape information around each point into the dynamic
programming matching process, resulting in more semantically meaningful
alignment results, i.e., points with similar local shapes tend to
be matched while those with dissimilar neighborhoods are unlikely
to be matched. shapeDTW consists of two steps: (1) represent each
temporal point by some shape descriptor; and (2) align two sequences
of descriptors by DTW. We first introduce the shapeDTW alignment framework,
and in the next section, we introduce several local shape descriptors.

Given a univariate time series $\mathcal{T}=(t_{1},t_{2},...,t_{L})^{T},\mathcal{T}\in\mathcal{R}^{L}$,
shapeDTW begins by representing each temporal point $t_{i}$ by a
shape descriptor $d_{i}\in\mathcal{R}^{m}$, which encodes structural
information of temporal neighborhoods around $t_{i}$, in this way,
the original real value sequence $\mathcal{T}=(t_{1},t_{2},...,t_{L})^{T}$
is converted to a sequence of shape descriptors of the same length,
i.e., $\mathbf{d}=(d_{1},d_{2},...,d_{L})^{T},\mathbf{d}\in\mathcal{R}^{L\times m}$.
shapeDTW then aligns the transformed multivariate descriptor sequences
$\mathbf{d}$ by DTW, and at last the alignment path between descriptor
sequences is transferred to the original univariate time series sequences.
We give implementation details of shapeDTW:

Given a univariate time series of length $L$, e.g.,$\mathcal{T}=(t_{1},t_{2},...,t_{L})^{T}$,
we first extract a subsequence $s_{i}$ of length $l$ from each temporal
point $t_{i}$. The subsequence $s_{i}$ is centered on $t_{i}$,
with its length $l$ typically much smaller than $L$($l\ll L$).
Note we have to pad both ends of $\mathcal{T}$ by $\lfloor\frac{l}{2}\rfloor$
with duplicates of $t_{1}(t_{L})$ to make subsequences sampled at
endpoints well defined. Now we obtain a sequence of subsequences,
i.e., $\mathcal{S}=(s_{1},s_{2},...,s_{L})^{T},\, s_{i}\in\mathcal{R}^{l}$,
with $s_{i}$ corresponding to the temporal point $t_{i}$. Next,
we design shape descriptors to express subsequences, under the goal
that similarly-shaped subsequences have similar descriptors while
differently-shaped subsequences have distinct descriptors. The shape
descriptor of subsequence $s_{i}$ naturally encodes local structural
information around the temporal point $t_{i}$, and is named as shape
descriptor of the temporal point $t_{i}$ as well. Designing a shape descriptor boils
down to designing a mapping function $\mathcal{F}(\cdot)$, which
maps subsequence $s_{i}\in\mathcal{R}^{l}$ to shape descriptor $d_{i}\in\mathcal{R}^{m}$,
i.e., $d_{i}=\mathcal{F}(s_{i})$, so that similarity between descriptors
can be measured simply with the Euclidean distance. Different mapping
functions define different shape descriptors, and one straightforward
mapping function is the identity function $\mbox{\ensuremath{\mathcal{I}}(\ensuremath{\cdot})}$,
in this case, $d_{i}=\mathcal{I}(s_{i})=s_{i}$, i.e., subsequence
itself acts as local shape descriptor. Given a shape descriptor computation
function $\mathcal{F}(\cdot)$, we convert the subsequence sequence
$\mathcal{S}$ to a descriptor sequence $\mathbf{d}=(d_{1},d_{2},...,d_{L})^{T}\, d_{i}\in\mathcal{R}^{m}$,
i.e., $\mathbf{d}=\mathcal{F}(\mathcal{S})=\left(\mathcal{F}(s_{1}),\mathcal{F}(s_{2}),...,\mathcal{F}(s_{L})\right)^{T}$.
At last, we use DTW to align two descriptor sequences and transfer
the warping path to the original univariate time series.

Given two univariate time series $\mathcal{P}=(p_{1},p_{2},...,p_{\mathcal{L_{P}}})^{T},\mathcal{P}\in\mathcal{R}^{\mathcal{L_{P}}}$
and $\mathcal{Q}=(q_{1},q_{2},...,q_{\mathcal{L_{Q}}})^{T},\mathcal{Q}\in\mathcal{R}^{\mathcal{L_{Q}}}$,
let $\mathbf{d}^{\mathcal{P}}=(d_{1}^{\mathcal{P}},d_{2}^{\mathcal{P}},...,d_{\mathcal{L_{P}}}^{\mathcal{P}})^{T},\, d_{i}^{\mathcal{P}}\in\mathcal{R}^{m},\,\mathbf{d}^{\mathcal{P}}\in\mathcal{R}^{\mathcal{L_{P}}\times m}$
and $\mathbf{d}^{\mathcal{Q}}=(d_{1}^{\mathcal{Q}},d_{2}^{\mathcal{Q}},...,d_{\mathcal{L_{Q}}}^{\mathcal{Q}})^{T},\, d_{i}^{\mathcal{Q}}\in\mathcal{R}^{m},\,\mathbf{d}^{\mathcal{Q}}\in\mathcal{R}^{\mathcal{L_{Q}}\times m}$
be their shape descriptor sequences respectively, shapeDTW alignment
is equivalent to solving the optimization problem:

\vspace{-12pt}

\begin{equation}
{\arg\min}_{l,\,\mathcal{\tilde{W}^{P}}\in\{0,1\}^{l\times\mathcal{L_{P}}},\mathcal{\tilde{W}^{Q}}\in\{0,1\}^{l\times\mathcal{L_{Q}}}}\parallel\mathcal{\tilde{W}^{P}}\cdot\mathcal{\mathtt{d}^{P}}-\mathcal{\tilde{W}^{Q}}\cdot\mathtt{d}^{\mathcal{Q}}\parallel_{1,2}\label{eq:sDTW}
\end{equation}

Where $\mathcal{\tilde{W}^{P}}$ and $\mathcal{\tilde{W}^{Q}}$ are
warping matrices of $\mathbf{d}^{\mathcal{P}}$ and $\mathbf{d}^{\mathcal{Q}}$,
and $\parallel\cdot\parallel_{1,2}$ is the $\ell_{1}/\ell_{2}$-norm
of matrix, i.e., $\parallel\mathcal{M}_{p\times n}\parallel_{1,2}=\sum_{i=1}^{p}\parallel\mathcal{M}_{i}\parallel_{2}$,
where $\mathcal{M}_{i}$ is the $i^{th}$ row of matrix $\mathcal{M}$.
Program \ref{eq:sDTW} is a multivariate time series alignment problem,
and can be effectively solved by dynamic programming in time $\mathcal{O}(\mbox{\ensuremath{\mathcal{L_{P}}\times}\ensuremath{\ensuremath{\mathcal{L_{Q}}}}})$.
The key difference between DTW and shapeDTW is that: DTW measures
similarities between $p_{i}$ and $q_{j}$ by their Euclidean distance
$|p_{i}-q_{j}|$, while shapeDTW uses the Euclidean distance between
their shape descriptors, i.e., $\parallel d_{i}^{\mathcal{P}}-d_{j}^{\mathcal{Q}}\parallel_{2}$,
as the similarity measure. shapeDTW essentially handles local non-linear
warping, since it is inherently DTW, and, on the other hand, it prefers
matching points with similar neighborhood structures to points with
similar values. shapeDTW algorithm is described in Algo.\ref{alg:shape-DTW}.

\begin{algorithm}
\caption{\label{alg:shape-DTW}shape Dynamic Time Warping}

$\textbf{Inputs:}$ univariate time series $\mathcal{P}\in\mathcal{R}^{\mathcal{L_{P}}}$
and $\mathcal{Q}\in\mathcal{R^{L_{Q}}}$; subsequence length $l$;
shape descriptor function $\mathcal{F}$

$\textbf{shapeDTW:}$

1. Sample subsequences: $\mathcal{S^{P}}\leftarrow\mathcal{P},\,\mathcal{S^{Q}}\leftarrow\mathcal{Q}$;

2. Encode subsequences by shape descriptors:

$\quad\mathtt{d}^{\mathcal{P}}\leftarrow\mathcal{F}(\mathcal{S^{P}}),\,\mathtt{d}^{\mathcal{Q}}\leftarrow\mathcal{F}(\mathcal{S^{Q}})$;

3. Align descriptor sequences $\mathtt{d}^{\mathcal{P}}$ and $\mathtt{d}^{\mathcal{Q}}$
by DTW.

$\textbf{Outputs:}$

warping matrices: $\tilde{W}_{*}^{\mathcal{P}}$ and $\tilde{W}_{*}^{\mathcal{Q}}$;

shapeDTW distance: $\parallel\tilde{W}_{*}^{\mathcal{P}}\cdot\mathcal{\mathtt{d}^{P}}-\tilde{W}_{*}^{\mathcal{Q}}\cdot\mathtt{d}^{\mathcal{Q}}\parallel_{1,2}$ 
\end{algorithm}

\section{\label{sec:shape-descriptors}Shape descriptors}

shapeDTW provides a generic alignment framework, and users can design
shape descriptors adapted to their domain data characteristics and
feed them into shapeDTW for alignments. Here we introduce several
general shape descriptors, each of which maps a subsequence $s_{i}$
to a vector representation $d_{i}$, i.e., $d_{i}=\mathcal{F}(s_{i})$.


The length $l$ of subsequences defines the size of neighborhoods
around temporal points. When $l=1$, no neighborhood information is
taken into account. With increasing $l$, larger neighborhoods are
considered, and in the extreme case when $l=L$ ($L$ is the length
of the time series), subsequences sampled from different temporal
points become the same, i.e., the whole time series, in which case,
shape descriptors of different points resemble each other too much,
making temporal points less identifiable by shape descriptors. In
practice, $l$ is set to some appropriate value. But in this section,
we first let $l$ be any positive integers ($l\ge1$), which does
not affect the definition of shape descriptors. In Sec.\ref{sec:Experimental-validation},
we will experimentally explore the sensitivity of NN-shapeDTW to the choice of $l$.

\subsection{Raw-Subsequence}

Raw subsequence $s_{i}$ sampled around point $t_{i}$ can be directly
used as the shape descriptor of $t_{i}$, i.e., $d_{i}=\mathcal{I}(s_{i})=s_{i}$,
where $\mathcal{I}(\cdot)$ is the identity function. Although simple,
it inherently captures the local subsequence shape and helps to disambiguate
points with similar values but different local shapes.

\subsection{PAA}

Piecewise aggregate approximation (PAA) is introduced in \cite{keogh2001dimensionality,yi2000fast}
to approximate time series. Here we use it to approximate subsequences.
Given a $l$-dimensional subsequence $s_{i}$, it is divided into
$m$ ($m\le l$) equal-lengthed intervals, the mean value of temporal
points falling within each interval is calculated and a vector of
these mean values gives the approximation of $s_{i}$ and is used
as the shape descriptor $d_{i}$ of $s_{i}$, i.e., $\mathcal{F}(\cdot)=PAA,\: d_{i}=PAA(s_{i})$.

\subsection{DWT}

Discrete Wavelet Transform (DWT) is another widely used technique
to approximate time series instances. Again, here we use DWT to approximate
subsequences. Concretely, we use a Haar wavelet basis to decompose
each subsequence $s_{i}$ into 3 levels. The detail wavelet coefficients
of all three levels and the approximation coefficients of the third
level are concatenated to form the approximation, which is used the
shape descriptor $d_{i}$ of $s_{i}$, i.e., $\mathcal{F}(\cdot)=DWT,\: d_{i}=DWT(s_{i})$.

\subsection{Slope}

All the above three shape descriptors encode local shape information
inherently. However, they are not invariant to y-shift, to be concrete,
given two subsequences $p,q$ of exactly the same shape, but $p$
is a y-shifted relative to $q$, e.g., $p=q+\Delta\cdot\textbf{1}$,
where $\Delta$ is the magnitude of y-shift, then their shape descriptors
under $\textit{Raw-Subsequence}$, $\textit{PAA}$ and $\textit{DWT}$
differ approximately by $\Delta$ as well, i.e., $d(p)\approx d(q)+\Delta\cdot\textbf{1}$.
Although magnitudes do help time series classification, it is also
desirable that similarly-shaped subsequences have similar descriptors.
Here we further exploit three shape descriptors in experiments, $\textit{Slope}$, $\textit{Derivative}$ and $\textit{HOG1D}$,
which are invariant to y-shift.

Slope is extracted as a feature and used in time series classification
in \cite{baydogan2013bag,deng2013time}. Here we use it to represent
subsequences. Given a $l$-dimensional subsequence $s_{i}$, it is
divided into $m$ ($m\le l$) equal-lengthed intervals. Within each
interval, we employ the total least square (TLS) line fitting approach
\cite{forsyth2003modern} to fit a line according to points falling
within that interval. By concatenating the slopes of the fitted lines
from all intervals, we obtain a $m$-dimensional vector representation,
which is the slope representation of $s_{i}$, i.e., $\mathcal{F}(\cdot)=Slope,\: d_{i}=Slope(s_{i})$.

\subsection{Derivative}

Similar to $\textit{Slope}$, $\textit{Derivative}$ is y-shift invariant
if it is used to represent shapes. Given a subsequence $s$, its first-order
derivative sequence is $s^{\prime}$, where $s^{\prime}$ is the first
order derivative according to time $t$. To keep consistent with derivatives
used in derivative Dynamic Time Warping \cite{keogh2001derivative}
(dDTW), we follow their formula to compute numeric derivatives.

\subsection{HOG1D}

HOG1D is introduced in \cite{zhao2015classifying} to represent 1D time series sequences.
It inherits key concepts from the histogram of oriented gradients (HOG) descriptor \cite{dalal2005histograms}, and uses 
concatenated gradient histograms to represent shapes of temporal sequences. Similarly to $\textit{Slope}$ and $\textit{Derivative}$ descriptors, $\textit{HOG1D}$ is invariant to y-shift as well.

In experiments, we divide a subsequence into 2 non-overlapping intervals, compute gradient histograms (under 8 bins) in each interval and concatenate two histograms as the HOG1D descriptor (a 16D vector) of that subsequence. We refer interested readers to \cite{zhao2015classifying} for computation details of HOG1D. We have to emphasize that: in \cite{zhao2015classifying}, the authors introduce a global scaling factor $\sigma$ and tune it using all training sequences; but here, we fix $\sigma$ to be 0.1 in all our experiments, therefore, HOG1D computation on one subsequence takes only linear time $\mathcal{O}(l)$, where $l$ is the length of that subsequence. See our published code for details.

\subsection{Compound shape descriptors}

Shape descriptors, like $\textit{HOG1D}$, $\textit{Slope}$ and $\textit{Derivative}$,
are invariant to y-shift. However, in the application of matching
two subsequences, y-magnitudes may sometimes be important cues as
well, e.g., DTW relies on point-wise magnitudes for alignments. Shape
descriptors, like $\textit{Raw-Subsequence}$, $\textit{PAA}$ and
$\textit{DWT}$, encode magnitude information, thus they complement
y-shift invariant descriptors. By fusing pure-shape capturing and
magnitude-aware descriptors, the compound descriptor has the potential
to become more discriminative of subsequences. In the experiments,
we generate compound descriptors by concatenating two complementary
descriptors, i.e., $d=(d_{A},\gamma d_{B})$, where $\gamma$ is a
weighting factor to balance two simple descriptors, and $d$ is the
generated compound descriptor.

\section{Alignment quality evaluation}

\begin{figure}[htbp!]
\begin{centering}
\includegraphics[width=0.35\textwidth]{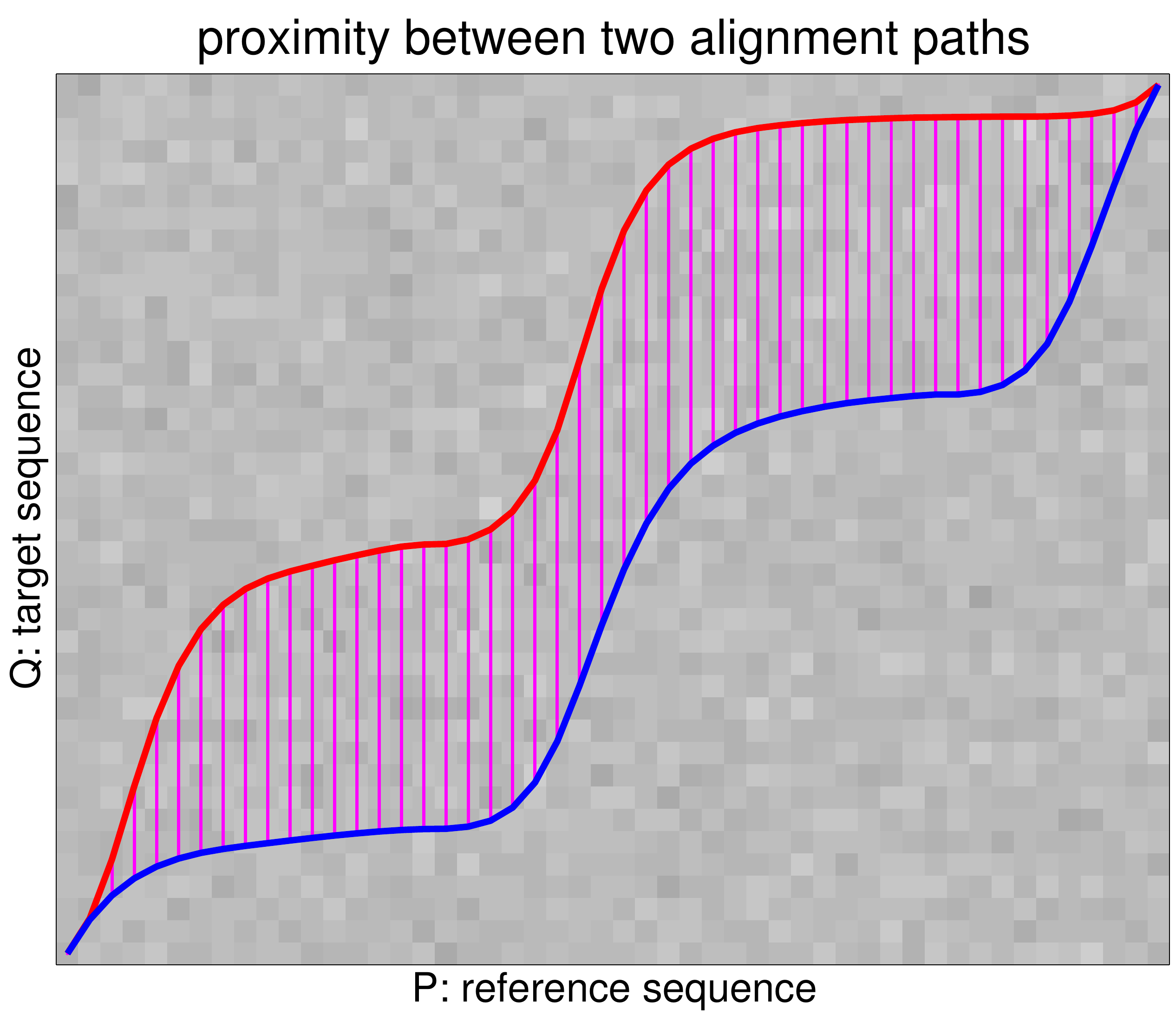} 
\par\end{centering}

\caption{\label{fig:fig-align-proximity} ``\textit{Mean absolute deviation}'', which
measures the proximity between alignment paths. The red and blue curves
are two alignment paths between sequences $\mathcal{P}$ and $\mathcal{Q}$,
and ``\textit{Mean absolute deviation}'' between these two paths is defined as:
the area of the slashed region divided by the length of the reference
sequence $\mathcal{P}$.}
\end{figure}

Here we adopt the ``\textit{mean absolute deviation}'' measure used
in the audio literature \cite{kirchhoff2011evaluation} to quantify
the proximity between two alignment paths. ``\textit{Mean absolute deviation}''
is defined as the mean distance between two alignment paths, which
is positively proportional to the area between two paths. Intuitively,
two spatially proximate paths have small between-areas, therefore
low ``\textit{Mean absolute deviation}''. Formally, given a reference
sequence $\mathcal{P}$, a target sequence $\mathcal{Q}$ and two
alignment paths $\alpha,\beta$ between them, the \textit{Mean absolute deviation}
between $\alpha$ and $\beta$ is calculate as: $\delta(\alpha,\beta)=\mathcal{A}(\alpha,\beta)/\mathcal{L_{P}}$,
where $\mathcal{A}(\alpha,\beta)$ is the area between $\alpha$ and
$\beta$ and $\mathcal{L_{P}}$ is the length of the reference sequence
$\mathcal{P}$. Fig. \ref{fig:fig-align-proximity} shows two alignment
paths $\alpha,\beta$, blue and red curves, between $\mathcal{P}$
and $\mathcal{Q}$. $\mathcal{A}(\alpha,\beta)$ is the area of the
slashed region, and in practice, it is computed by counting the number
of cells falling within it. Here a cell $(i,j)$ refers to the position
$(i,j)$ in the pairwise distance matrix $\mathcal{D}(\mathcal{P},\mathcal{Q})\in\mathcal{R^{L_{P}\times L_{Q}}}$
between $\mathcal{P}$ and $\mathcal{Q}$.

\section{\label{sec:Experimental-validation}Experimental validation}

We test shapeDTW for sequence alignment and time series classification
extensively on 84 UCR time series datasets \cite{UCRArchive} and
the Bach10 dataset \cite{duan2011soundprism}. For sequence alignment,
we compare shapeDTW against DTW and its other variants both qualitatively
and quantitatively: specifically, we first visually compare alignment
results returned by shapeDTW and DTW (and its variants), and then
quantify their alignment path qualities on both synthetic and real
data. Concretely, we simulate aligned pairs by artificially scaling
and stretching original time series sequences, align those pairs by
shapeDTW and DTW (and its variants), and then evaluate the alignment
paths against the ground-truth alignments. We further evaluate the
alignment performances of shapeDTW and DTW (and its variants) on audio
signals, which have the ground-truth point-to-point alignments. For
time series classification, since it is widely recognized that the
nearest neighbor classifier with the distance measure DTW (NN-DTW)
is very effective and is hard to beaten \cite{wang2013experimental,bagnall2014experimental},
we use the nearest neighbor classifier as well to test the effectiveness
of shapeDTW (NN-shapeDTW), and compare NN-shapeDTW against NN-DTW.
We further compare NN-shapeDTW against six other state-of-the-art
classification algorithms in the supplementary materials.

\subsection{Sequence alignment}

We evaluate sequence alignments qualitatively in Sec. \ref{sub:Qualitative-alignment-assessment}
and quantitatively in Sec. \ref{sub:Simulated-sequence-pair-alignmen}
and Sec. \ref{sub:MIDI-to-audio-alignment}. We compare shapeDTW against
DTW, derivative Dynamic Time Warping (dDTW) \cite{keogh2001derivative}
and weighted Dynamic Time Warping (wDTW)\cite{jeong2011weighted}.
dDTW first computes derivative sequences, and then aligns them by
DTW. wDTW uses a weighted $\ell_{2}$ distance, instead of the regular
$\ell_{2}$ distance, to compute distances between points, and the
weight accounts for the phase differences between points. wDTW is
essentially a DTW algorithm. Here, both dDTW and wDTW are variants
of the original DTW. Before the evaluation, we briefly introduce some
popular step patterns in DTW.

\subsubsection{Step pattern in DTW}

\begin{figure}[htbp!]
\begin{center}
\includegraphics[width=0.5\textwidth]{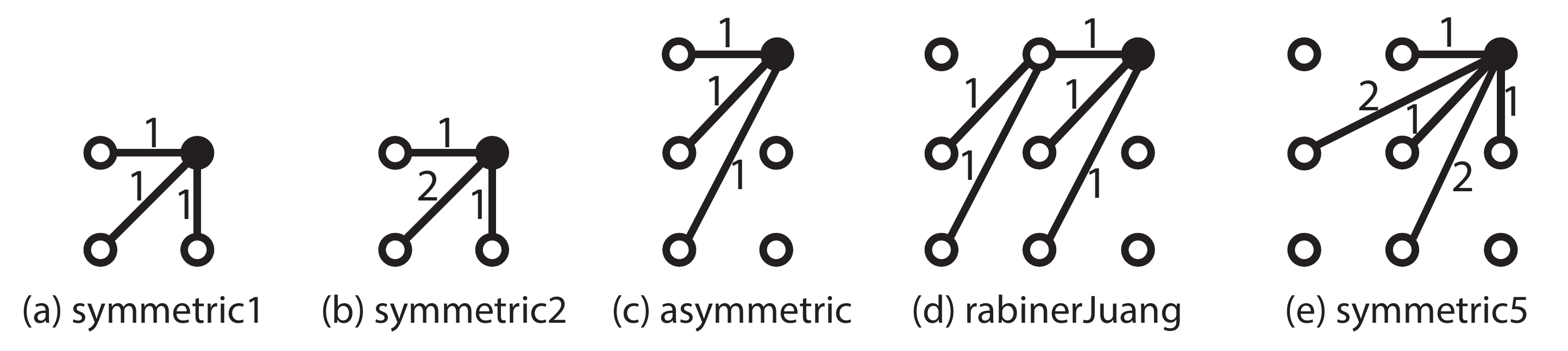}
\end{center}
\caption{\label{fig:Five-step-patterns.}Five step patterns. Numbers on transitions
indicate the multiplicative weight for the local distance $d(i,j)$.
Step-pattern (a) ``symmetric1'' is the default step pattern for DTW
and (b) gives more penalties to the diagonal directions, such that
the warping favors stair-stepping paths. Step patterns (a) and (b)
obtain a continuous warping path, while step patterns (c), (d) and
(e) may result in skipping elements, i.e., some temporal points from
one sequence are not matched to any points from the other sequence,
and vice verse. }
\end{figure}

Step pattern in DTW defines the allowed transitions between matched
pairs, and the corresponding weights. In both Program. \ref{eq:DTW} (DTW)
and Program. \ref{eq:sDTW} (shapeDTW), we use the default step pattern,
whose recursion formula is $D(i,j)=d(i,j)+\min\{D(i-1,j-1),\, D(i,j-1),\, D(i-1,j)\}$.
In the following alignment experiments, we try other well-known step
patterns as well, and we follow the naming convention in \cite{giorgino2009computing}
to name these step-patterns. Five popular step-patterns, ``symmetric1'',
``symmetric2'', ``symmetric5'', ``asymmetric'' and ``rabinerJuang'', are listed
in Fig. \ref{fig:Five-step-patterns.}. Step-pattern (a), ``symmetric1'',
is the one used by shapeDTW in all the following alignment and classification
experiments, and we will not explicitly mention that in following
texts.

\subsubsection{\label{sub:Qualitative-alignment-assessment}Qualitative alignment
assessment}

We plot alignment results by shapeDTW and DTW/dDTW, and
evaluate them visually. shapeDTW under 5 shape descriptors,
$\textit{Raw-Subsequence}$, $\textit{PAA}$, $\textit{DWT}$, $\textit{Derivative}$
and $\textit{HOG1D}$, obtains similar alignment results, here we
choose $\textit{Derivative}$ as a representative to report results,
with the subsequence length set to be 30. Here, shapeDTW, DTW and
dDTW all use step pattern (a) in Fig. \ref{fig:Five-step-patterns.}.

\begin{figure*}[htbp!]
\begin{centering}
\includegraphics[width=1.0\textwidth]{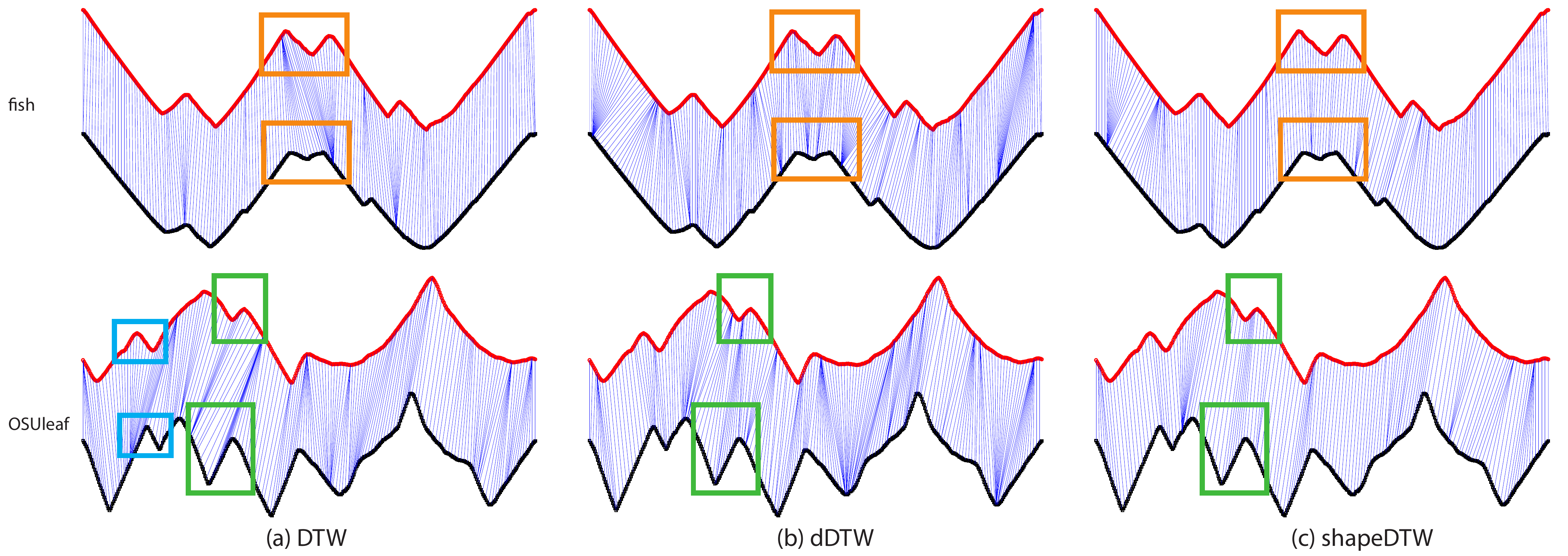} 
\par\end{centering}

\caption{\label{fig:fig-rich-alignment} Alignments between time series with
rich local features. Time series at the top and bottom row are from
``Fish''(train-165, test-1) and ``OSUleaf''(test-114, test-134) datasets
respectively. In each pair of time series, temporal points with similar
local structures are boxed out by rectangles. Perceptually, shapeDTW
aligns these corresponding points better than both DTW and dDTW.}
\end{figure*}

Time series with rich local features: time series with
rich local features, such as those in the ``OSUleaf'' dataset (bottom
row in Fig.\ref{fig:fig-rich-alignment}), have many bumps and valleys;
DTW becomes quite brittle to align such sequences, since it matches
two points based on their single-point y-magnitudes. Because single
magnitude value does not incorporate local neighborhood information,
it is hard for DTW to discriminate a peak point $\mathtt{p}$ from
a valley point $\mathtt{v}$ with the same magnitude, although $\mathtt{p}$
and $\mathtt{v}$ have dramatically different local shapes. dDTW bears
similar weakness as DTW, since it matches points bases on their derivative
differences and does not take local neighborhood into consideration
either. On the contrary, shapeDTW distinguishes peaks from valleys
easily by their highly different local shape descriptors. Since shapeDTW
takes both non-linear warping and local shapes into account, it gives
more perceptually interpretable and semantically sensible alignments
than DTW (dDTW). Some typical alignment results of time series from
feature rich datasets ``OSUleaf'' and ``Fish'' are shown in Fig.\ref{fig:fig-rich-alignment}.

\subsubsection{\label{sub:Simulated-sequence-pair-alignmen}Simulated sequence-pair
alignment}

We simulate aligned sequence pairs by scaling and stretching original
time series. Then we run shapeDTW and DTW (and its variants) to align
the simulated pairs, and compare their alignment paths against the
ground-truth. In this section, shapeDTW is run under the fixed settings:
(1) fix the subsequence length to be 30, (2) use $\textit{Derivative}$
as the shape descriptor and (3) use ``symmetric1'' as the step-pattern.

\noindent $\textbf{Aligned-pairs simulation algorithm}:$ concretely, given
a time series $\mathcal{T}$ of length $L$,
we simulate a new time series by locally scaling and stretching $\mathcal{T}$.
The simulation consists of two sequential steps: (1) scaling: scale
$\mathcal{T}$ point-wisely, resulting in a new time series $\mathcal{\hat{T}}=\mathcal{T}\otimes S$,
where $S$ is a positive scale vector with the same length as $\mathcal{T}$,
and $\otimes$ is a point-wise multiplication operator; (2) stretching:
randomly choose $\alpha$ percent of temporal points from $\mathcal{\hat{T}}$,
stretch each point by a random length $\tau$ and result in a new
time series $\mathcal{T^{\prime}}$. $\mathcal{T^{\prime}}$ and $\mathcal{T}$
are a simulated alignment pair, with the ground-truth alignment known
from the simulation process. The simulation algorithm is described
in Alg. \ref{alg:simulate-alignment-pairs}.

One caveat we have to pay attention to is that: scaling an input time
series by a random scale vector can make the resulting time series
perceptually quite different from the original one, such that simulated
alignment pairs make little sense. Therefore, in practice, a scale
vector $S$ should be smooth, i.e., adjacent elements in $S$ cannot
be random, instead, they should be similar in magnitude, making adjacent
temporal points from the original time series be scaled by a similar
amount. In experiments, we first use a random process, which is similar
to Brownian motion, to initialize scale vectors, and then recursively
smooth it. The scale vector generation algorithm is shown in Alg.
\ref{alg:simulate-alignment-pairs}. As seen, adjacent scales are
initialized to be differed by at most 1 (i.e., $s(t+1)=s(t)+sin\left(\pi\times randn\right)$),
such that the first order derivatives are bounded and initialized
scale vectors do not change abruptly. Initialized scale vectors usually
have local bumps, and we further recursively utilize cumulative summation
and sine-squashing, as described in the algorithm, to smooth the scale
vectors. Finally, the smoothed scale vectors are linearly squashed
into a positive range $[a\: b]$.

After non-uniformly scaling an input time series by a scale vector,
we obtain a scale-transformed new sequence, and then we randomly pick
$\alpha$ percent of points from the new sequence and stretch each
of them by some random amount $\tau$. Stretching at point $p$ by
some amount $\tau$ is to duplicate $p$ by $\tau$ times.

\begin{algorithm}
\caption{\label{alg:simulate-alignment-pairs}simulate alignment pairs}

$\textbf{Simulate an alignment pair}:$

$\textbf{Inputs:}$ a time series instance $\mathcal{T}$; scale vector
range $[a\: b]$, smoothing iterations $\Gamma$; stretching percentage
$\alpha$, stretching amount $\tau$

1. simulate a scale vector $S$;

2. scale $\mathcal{T}$ point-wisely, $\mathcal{\hat{T}}\leftarrow\mathcal{T}\otimes S$;

3. stretching $\alpha$ percent of points from $\mathcal{\hat{T}}$
by a random amount $\tau$, resulting in a simulated time series $\mathcal{T^{\prime}}$.

$\textbf{Outputs:}$$\mathcal{T^{\prime}}$

$\textbf{Simulate a scale vector:}$

$\textbf{Inputs:}$ length $L$, iteration $\Gamma$, range $[a\: b]$

1.$\textit{Initialize}$:

$\begin{cases}
s(1)=randn\\
s(t+1)=s(t)+sin\left(\pi\times randn\right),t\in\{1,2,...,L-1\}
\end{cases}$

2.$\textit{smoothing}$:

$\textit{while}$ iteration $<$ $\Gamma$

a. set the cumulative sum up to $t$ as the scale at $t$:

$\begin{cases}
s(1)\leftarrow s(1)\\
s(t+1)\leftarrow s(t+1)+s(t),t\in\{1,2,...,L-1\}
\end{cases}$

b. squash scale at $t$ into the range $[-1,1]$:

$s(t)\leftarrow sin\left(s(t)\right)$, $t\in\{1,2,...,L\}$

$\textit{end}$

3. squash elements in the scale vector $S$ into range $[a\; b]$
by linear scaling.

$\textbf{Outputs:}$ a scale vector $S=\{s(1),s(2),...,s(L)\}$ 
\end{algorithm}

\begin{figure*}[htbp!]
\begin{centering}
\includegraphics[width=1.0\textwidth]{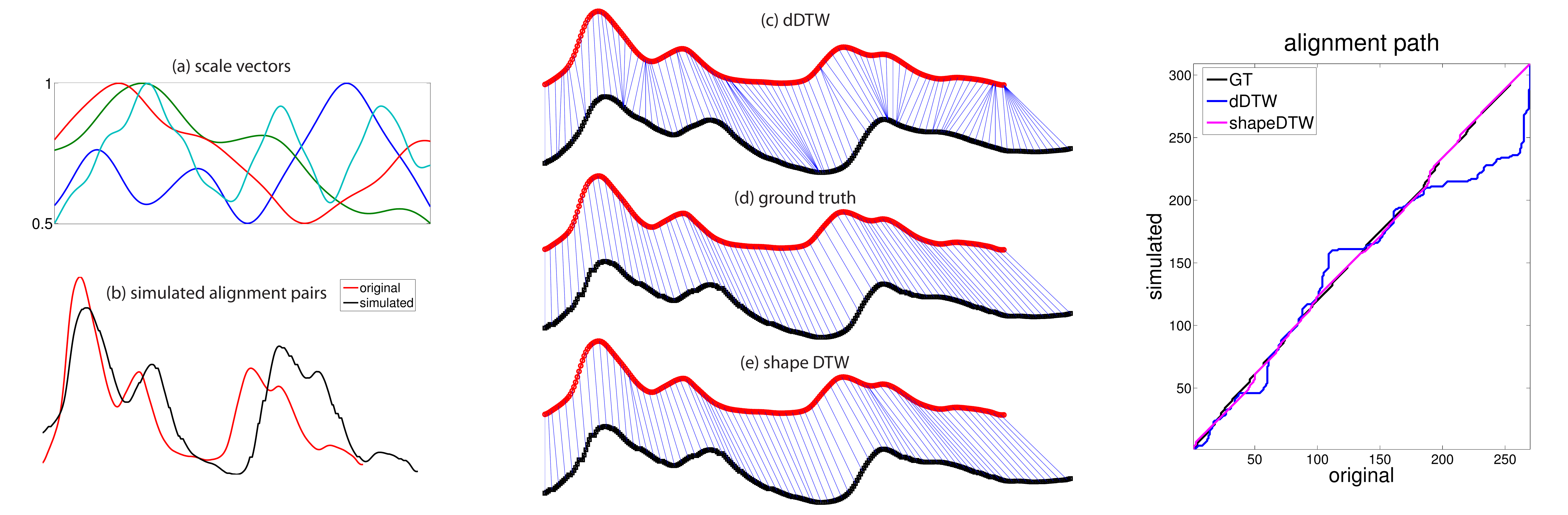} 
\par\end{centering}

\caption{\label{fig:fig-align-simulation} Alignments between simulated time
series pairs. (a) simulated scale vectors: they are smooth and squashed
to the range $[0.5\:1.0]$; (b) a simulated alignment pair: generated
by artificially scale and stretch the original time series; (c) dDTW
alignment: run dDTW to align the simulated pair; (d) ground truth
alignment; (e) shapeDTW alignment. The plot on the right shows alignment
paths of dDTW, shapeDTW and the ground-truth, visually the alignment
path of shapeDTW is closer to the ground-truth, and quantitatively,
shapeDTW has 1.1 alignment errors in terms of ``\textit{Mean Absolute Deviation}''
score, compared with 4.7 of dDTW.}
\end{figure*}

\noindent $\textbf{Aligned-pairs simulation}:$ using training data from each
UCR dataset as the original time series, we simulate their alignment
pairs by running Alg. \ref{alg:simulate-alignment-pairs}. Since there
are 27,136 training time series instances from 84 UCR datasets, we
simulate 27,136 aligned-pairs in total. We fix most simulation parameters
as follows: $[a\: b]=[0.5\:1]$, $\Gamma=5$, $\tau=\{1,2,3\}$, and
the stretching percentage $\alpha$ is the only flexible parameter
we will vary, e.g., when $\alpha=15\%$, each original input time
series is on average stretched by $30\%$ (in length). Typical scale
vectors and simulated alignment pairs are shown in Fig. \ref{fig:fig-align-simulation}.
The scale vectors are smooth and the simulated time series are both
scaled and stretched, compared with the original ones.

\noindent $\textbf{Alignment comparison between shapeDTW and DTWs}:$
we run shapeDTW and DTW/dDTW/wDTW to align simulated pairs, and compare
alignment paths against the ground-truth in terms of ``\textit{Mean Absolute Deviation}''
scores. DTW and dDTW are parameter-free, but wDTW has one tuning parameter
$g$ (see Eq. (3) in their paper), which controls the curvature of
the logistic weight function. However in the case of aligning two
sequences, $g$ is impossible to be tuned and should be pre-defined
by experiences. Here we fix $g$ to be 0.1, which is the approximate
mean value of the optimal $g$ in the original paper. For the purpose
of comparing the alignment qualities of different algorithms, we use
the default step pattern, (a) in Fig. \ref{fig:Five-step-patterns.},
for both shapeDTW and DTW/dDTW/wDTW, but we further evaluate effects
of different step-patterns in the following experiments.

\begin{figure}
\begin{centering}
\includegraphics[width=0.4\textwidth]{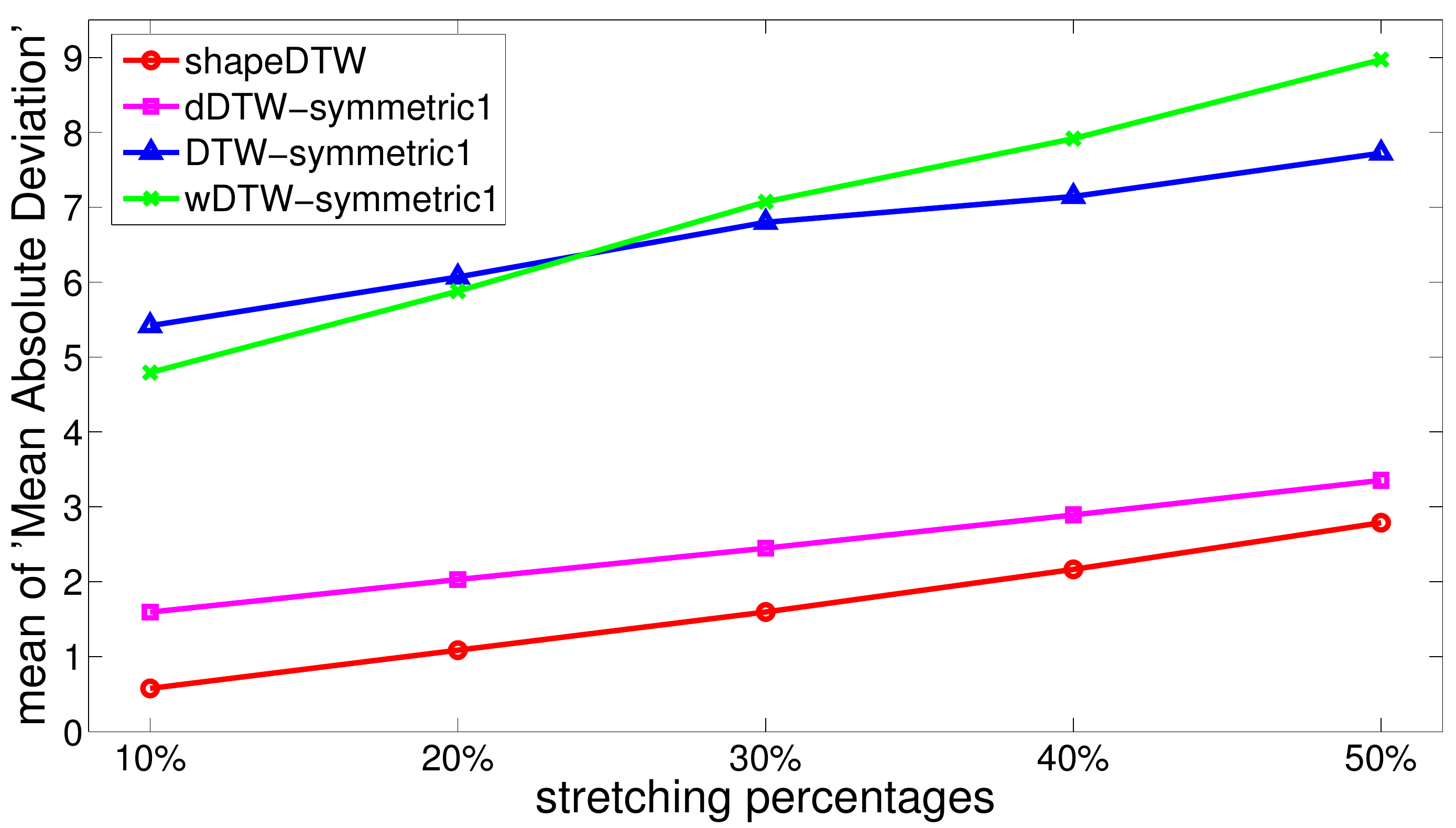} 
\par\end{centering}

\caption{\label{fig:Alignment-under-different} Alignment quality comparison
between shapeDTW and DTW/dDTW/wDTW, under the step pattern ``symmetric1''.
As seen, as the stretching amount increases, the alignment qualities
of both shapeDTW and DTW/dDTW/wDTW drop. However, shapeDTW consistently
achieves lower alignment errors under different stretching amounts,
compared with DTW, dDTW and wDTW.}
\end{figure}

We simulate alignment pairs by stretching raw time series by different
amounts, $10\%$, $20\%$, $30\%$, $40\%$ and $50\%$, and report
the alignment qualities of shapeDTW and DTW/dDTW/wDTW under each stretching
amount in terms of the mean of ``\textit{Mean Absolute Deivation}''
scores over 27,136 simulated pairs. The results are shown in Fig.
\ref{fig:Alignment-under-different}, which shows shapeDTW achieves
lower alignment errors than DTW / dDTW / wDTW over different stretching
amounts consistently. shapeDTW almost halves the alignment errors
achieved by dDTW, although dDTW already outperforms its two competitors,
DTW and wDTW, by a large margin.

\begin{figure*}[htbp!]
\begin{center}
\includegraphics[width=1.0\textwidth]{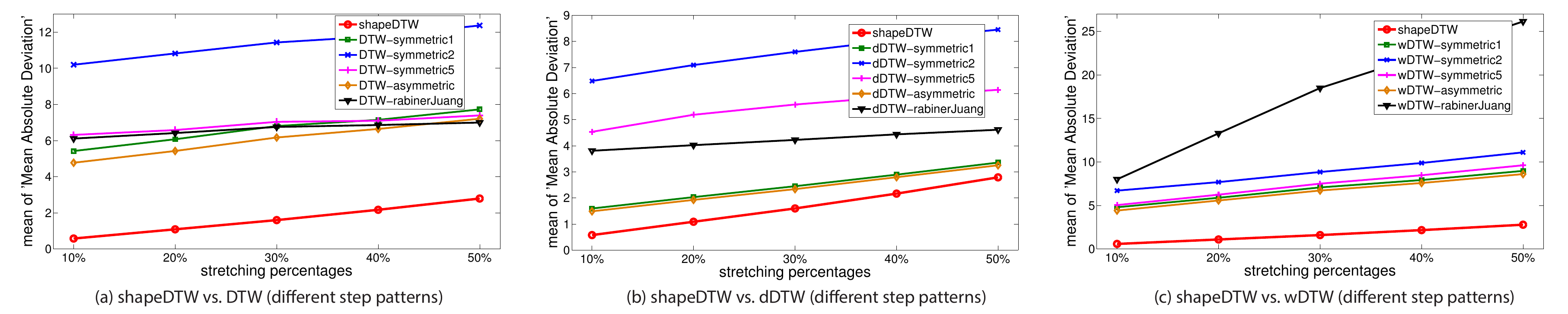}
\end{center}
\caption{\label{fig:different-steppattern-res}Align sequences under different
step patterns. We align sequence-pairs by DTW/dDTW/wDTW under 5 different
step patterns (Fig.~\ref{fig:Five-step-patterns.}), ``symmetric1'', ``symmetric2'',
``symmetric5'', ``asymmetric'' and ``rabinerJuang'', and compare their alignment
errors against those obtained by shapeDTW. As seen, different step
patterns usually reach different alignment results, which shows the
importance of choosing an appropriate step pattern adapted to the
application domain. In our case, ``asymmetric'' step pattern achieves
slightly lower errors than ``symmetric1'' step pattern (under DTW, wDTW
and dDTW), however, shapeDTW consistently wins DTW/dDTW/wDTW under
the best step pattern - ``asymmetric''. }

\end{figure*}

\noindent $\textbf{Effects of different step patterns}:$ choosing a suitable
step pattern is a traditionally way to improve sequence alignments,
and it usually needs domain knowledge to make the right choice. Here,
instead of choosing an optimal step pattern, we run DTW/dDTW/wDTW
under all 5 step patterns in Fig. \ref{fig:Five-step-patterns.} and
compare their alignment performances against shapeDTW. Similar as
the above experiments, we simulate aligned-pairs under different amounts
of stretches, report alignment errors under different step patterns
in terms of the mean of ``\textit{Mean Absolute Deivation}'' scores
over 27,136 simulated pairs, and plot the results in Fig. \ref{fig:different-steppattern-res}.
As seen, different step patterns obtain different alignment qualities,
and in our case, step patterns, ``symmetric1'' and ``asymmetric'', have
similar alignment performances and they reach lower alignment errors
than the other 3 step patterns. However, shapeDTW still wins DTW/dDTW/wDTW
(under ``symmetric1'' and ``asymmetric'' step-patterns) by some margin.

\begin{table*}[!htbp]
\global\long\def\tabincell#1#2{\begin{tabular}{@{}@{}}
 #2\end{tabular}}
 \centering %
 \resizebox{2.0\columnwidth}{!}{
\begin{tabular}{p{2.7cm}ccccp{3.2cm}cccc}
\toprule 
\multicolumn{10}{c}{{\normalsize{}{}{}{}{}}\textbf{\textit{\normalsize{}{}{}{}Mean Absolute
Deviation}}\textbf{\normalsize{}{}{}{} from the ground-truth alignments}{\normalsize{}{}{}{}}}\tabularnewline [-1pt]
\midrule 
{}  & \multicolumn{2}{c}{\textbf{mean}} & \multicolumn{2}{c}{\textbf{std.}} & {}  & \multicolumn{2}{c}{\textbf{mean}} & \multicolumn{2}{c}{\textbf{std.}}\tabularnewline [-1pt]
{\normalsize{}{}{}{datasets} }  & {\normalsize{}{}{}{shapeDTW} }  & {\normalsize{}{}{}{dDTW} }  & {\normalsize{}{}{}{shapeDTW} }  & {\normalsize{}{}{}{dDTW} }  & {\normalsize{}{}{}{datasets} }  & {\normalsize{}{}{}{shapeDTW} }  & {\normalsize{}{}{}{dDTW} }  & {\normalsize{}{}{}{shapeDTW} }  & {\normalsize{}{}{}{dDTW} }\tabularnewline [-1pt]
\midrule 
\small{50words} & \small{\textbf{1.49}} & \small{2.85} & \small{\textbf{1.03}}  & \small{2.14} & 	\small{MedicalImages} & \small{\textbf{0.93}} & \small{2.14} & \small{\textbf{0.66}}  & \small{2.05} \tabularnewline
\small{Adiac} & \small{\textbf{1.77}} & \small{5.73} & \small{\textbf{0.61}}  & \small{2.45} & 	\small{MiddlePhalanxOutlineAgeGroup} & \small{\textbf{0.47}} & \small{0.80} & \small{\textbf{0.18}}  & \small{0.30} \tabularnewline
\small{ArrowHead} & \small{\textbf{0.94}} & \small{1.70} & \small{\textbf{0.48}}  & \small{0.91} & 	\small{MiddlePhalanxOutlineCorrect} & \small{\textbf{0.46}} & \small{0.81} & \small{\textbf{0.18}}  & \small{0.27} \tabularnewline
\small{Beef} & \small{\textbf{0.85}} & \small{1.86} & \small{\textbf{0.22}}  & \small{0.83} & 	\small{MiddlePhalanxTW} & \small{\textbf{0.53}} & \small{0.91} & \small{\textbf{0.27}}  & \small{0.39} \tabularnewline
\small{BeetleFly} & \small{\textbf{0.69}} & \small{2.22} & \small{\textbf{0.16}}  & \small{0.80} & 	\small{MoteStrain} & \small{\textbf{0.78}} & \small{1.07} & \small{\textbf{0.31}}  & \small{0.85} \tabularnewline
\small{BirdChicken} & \small{\textbf{1.11}} & \small{2.35} & \small{\textbf{0.85}}  & \small{1.65} & 	\small{NonInvasiveFatalECG-Thorax1} & \small{\textbf{0.65}} & \small{0.72} & \small{\textbf{0.24}}  & \small{0.49} \tabularnewline
\small{Car} & \small{\textbf{1.83}} & \small{6.34} & \small{\textbf{1.74}}  & \small{3.21} & 	\small{NonInvasiveFatalECG-Thorax2} & \small{\textbf{0.80}} & \small{1.06} & \small{\textbf{0.51}}  & \small{0.89} \tabularnewline
\small{CBF} & \small{0.60} &  \small{\textbf{0.13}} & \small{0.28} & \small{\textbf{0.03}} & 	\small{OliveOil} & \small{\textbf{1.89}} & \small{3.90} & \small{0.79} & \small{\textbf{0.69}}   \tabularnewline
\small{ChlorineConcentration} & \small{0.64} & \small{\textbf{0.23}} & \small{\textbf{0.18}}  & \small{0.24} & 	\small{OSULeaf} & \small{\textbf{0.69}} & \small{1.92} & \small{\textbf{0.17}}  & \small{0.94} \tabularnewline
\small{CinC-ECG-torso} & \small{0.69} & \small{\textbf{0.67}} & \small{\textbf{0.33}}  & \small{0.92} & 	\small{PhalangesOutlinesCorrect} & \small{\textbf{0.62}} & \small{1.04} & \small{\textbf{0.29}}  & \small{0.49} \tabularnewline
\small{Coffee} & \small{\textbf{0.69}} & \small{1.36} & \small{\textbf{0.17}}  & \small{0.41} & 	\small{Phoneme} & \small{\textbf{0.69}} & \small{0.89} & \small{\textbf{0.52}}  & \small{5.37} \tabularnewline
\small{Computers} & \small{11.18} & \small{\textbf{10.73}} & \small{\textbf{12.62}}  & \small{13.10} & 	\small{Plane} & \small{\textbf{0.51}} & \small{1.44} & \small{\textbf{0.16}}  & \small{0.59} \tabularnewline
\small{Cricket-X} & \small{0.64} &  \small{\textbf{0.18}} & \small{0.17} & \small{\textbf{0.07}} & 	\small{ProximalPhalanxOutlineAgeGroup} & \small{\textbf{0.60}} & \small{1.15} & \small{\textbf{0.29}}  & \small{0.52} \tabularnewline
\small{Cricket-Y} & \small{0.65} &  \small{\textbf{0.19}} & \small{0.16} & \small{\textbf{0.07}} & 	\small{ProximalPhalanxOutlineCorrect} & \small{\textbf{0.61}} & \small{1.14} & \small{\textbf{0.29}}  & \small{0.49} \tabularnewline
\small{Cricket-Z} & \small{0.65} &  \small{\textbf{0.19}} & \small{0.15} & \small{\textbf{0.06}} & 	\small{ProximalPhalanxTW} & \small{\textbf{0.56}} & \small{1.08} & \small{\textbf{0.26}}  & \small{0.49} \tabularnewline
\small{DiatomSizeReduction} & \small{\textbf{2.21}} & \small{7.43} & \small{\textbf{1.15}}  & \small{2.82} & 	\small{RefrigerationDevices} & \small{1.28} & \small{\textbf{1.21}} & \small{\textbf{1.33}}  & \small{1.53}  \tabularnewline
\small{DistalPhalanxOutlineAgeGroup} & \small{\textbf{0.57}} & \small{0.88} & \small{\textbf{0.27}}  & \small{0.49} & 	\small{ScreenType} & \small{11.26} & \small{\textbf{11.00}} & \small{\textbf{11.29}}  & \small{11.70}  \tabularnewline
\small{DistalPhalanxOutlineCorrect} & \small{\textbf{0.57}} & \small{0.85} & \small{\textbf{0.30}}  & \small{0.49} & 	\small{ShapeletSim} & \small{0.60} &  \small{\textbf{0.20}} & \small{0.17} & \small{\textbf{0.10}} \tabularnewline
\small{DistalPhalanxTW} & \small{\textbf{0.60}} & \small{0.85} & \small{\textbf{0.25}}  & \small{0.44} & 	\small{ShapesAll} & \small{\textbf{1.13}} & \small{3.07} & \small{\textbf{0.79}}  & \small{2.65} \tabularnewline
\small{Earthquakes} & \small{0.97} & \small{\textbf{0.77}} & \small{\textbf{0.59}}  & \small{0.76} & 	\small{SmallKitchenAppliances} & \small{\textbf{15.77}} & \small{15.88} & \small{\textbf{11.67}}  & \small{12.16} \tabularnewline
\small{ECG200} & \small{0.61} &  \small{\textbf{0.22}} & \small{0.25} & \small{\textbf{0.15}} & 	\small{SonyAIBORobotSurface} & \small{0.63} &  \small{\textbf{0.18}} & \small{0.16} & \small{\textbf{0.11}} \tabularnewline
\small{ECG5000} & \small{0.67} &  \small{\textbf{0.24}} & \small{0.25} & \small{\textbf{0.13}} & 	\small{SonyAIBORobotSurfaceII} & \small{0.79} &  \small{\textbf{0.16}} & \small{0.28} & \small{\textbf{0.06}} \tabularnewline
\small{ECGFiveDays} & \small{0.78} &  \small{\textbf{0.26}} & \small{0.33} & \small{\textbf{0.15}} & 	\small{Strawberry} & \small{\textbf{0.71}} & \small{1.07} & \small{\textbf{0.17}}  & \small{0.47} \tabularnewline
\small{FaceAll} & \small{0.54} &  \small{\textbf{0.23}} & \small{0.19} & \small{\textbf{0.11}} & 	\small{SwedishLeaf} & \small{\textbf{0.65}} & \small{1.36} & \small{\textbf{0.29}}  & \small{1.25} \tabularnewline
\small{FaceFour} & \small{0.64} &  \small{\textbf{0.16}} & \small{0.10} & \small{\textbf{0.07}} & 	\small{Symbols} & \small{\textbf{2.20}} & \small{6.04} & \small{\textbf{1.73}}  & \small{4.17} \tabularnewline
\small{FacesUCR} & \small{0.55} &  \small{\textbf{0.25}} & \small{0.17} & \small{\textbf{0.12}} & 	\small{synthetic-control} & \small{0.54} &  \small{\textbf{0.12}} & \small{0.59} & \small{\textbf{0.07}} \tabularnewline
\small{FISH} & \small{\textbf{2.64}} & \small{10.56} & \small{\textbf{1.16}}  & \small{3.33} & 	\small{ToeSegmentation1} & \small{0.69} & \small{\textbf{0.36}} & \small{\textbf{0.17}}  & \small{0.19}  \tabularnewline
\small{FordA} & \small{\textbf{0.50}} & \small{0.59} & \small{\textbf{0.07}}  & \small{0.20} & 	\small{ToeSegmentation2} & \small{0.65} & \small{\textbf{0.45}} & \small{\textbf{0.22}}  & \small{0.42}  \tabularnewline
\small{FordB} & \small{\textbf{0.50}} & \small{0.64} & \small{\textbf{0.07}}  & \small{0.15} & 	\small{Trace} & \small{0.62} &  \small{\textbf{0.25}} & \small{0.24} & \small{\textbf{0.22}} \tabularnewline
\small{Gun-Point} & \small{\textbf{1.36}} & \small{5.95} & \small{\textbf{0.65}}  & \small{2.86} & 	\small{TwoLeadECG} & \small{0.78} & \small{\textbf{0.78}} & \small{\textbf{0.23}}  & \small{0.39}  \tabularnewline
\small{Ham} & \small{0.73} & \small{\textbf{0.63}} & \small{\textbf{0.13}}  & \small{0.29} & 	\small{Two-Patterns} & \small{0.63} &  \small{\textbf{0.35}} & \small{0.22} & \small{\textbf{0.17}} \tabularnewline
\small{HandOutlines} & \small{\textbf{7.08}} & \small{22.92} & \small{\textbf{6.86}}  & \small{6.92} & 	\small{UWaveGestureLibraryAll} & \small{\textbf{1.13}} & \small{2.45} & \small{\textbf{0.79}}  & \small{2.57} \tabularnewline
\small{Haptics} & \small{\textbf{1.37}} & \small{2.00} & \small{\textbf{0.70}}  & \small{1.19} & 	\small{uWaveGestureLibrary-X} & \small{\textbf{1.44}} & \small{3.69} & \small{\textbf{1.62}}  & \small{4.02} \tabularnewline
\small{Herring} & \small{\textbf{0.97}} & \small{3.44} & \small{\textbf{0.40}}  & \small{1.45} & 	\small{uWaveGestureLibrary-Y} & \small{\textbf{1.39}} & \small{3.98} & \small{\textbf{1.33}}  & \small{4.48} \tabularnewline
\small{InlineSkate} & \small{0.68} &  \small{\textbf{0.19}} & \small{0.16} & \small{\textbf{0.09}} & 	\small{uWaveGestureLibrary-Z} & \small{\textbf{1.52}} & \small{4.07} & \small{\textbf{1.52}}  & \small{4.44} \tabularnewline
\small{InsectWingbeatSound} & \small{\textbf{1.09}} & \small{3.06} & \small{\textbf{0.64}}  & \small{3.98} & 	\small{wafer} & \small{\textbf{1.08}} & \small{5.26} & \small{\textbf{0.58}}  & \small{3.36} \tabularnewline
\small{ItalyPowerDemand} & \small{0.60} &  \small{\textbf{0.25}} & \small{0.39} & \small{\textbf{0.16}} & 	\small{Wine} & \small{\textbf{1.25}} & \small{2.00} & \small{\textbf{0.26}}  & \small{0.45} \tabularnewline
\small{LargeKitchenAppliances} & \small{\textbf{19.88}} & \small{20.38} & \small{\textbf{12.03}}  & \small{15.98} & 	\small{WordsSynonyms} & \small{\textbf{1.57}} & \small{2.89} & \small{\textbf{1.23}}  & \small{2.29} \tabularnewline
\small{Lighting2} & \small{\textbf{1.71}} & \small{3.20} & \small{\textbf{0.58}}  & \small{2.72} & 	\small{WordSynonyms} & \small{\textbf{1.48}} & \small{2.92} & \small{\textbf{1.04}}  & \small{2.50} \tabularnewline
\small{Lighting7} & \small{\textbf{1.23}} & \small{1.90} & \small{\textbf{0.40}}  & \small{1.68} & 	\small{Worms} & \small{\textbf{0.65}} & \small{1.00} & \small{\textbf{0.12}}  & \small{3.48} \tabularnewline
\small{MALLAT} & \small{\textbf{2.45}} & \small{3.98} & \small{\textbf{0.50}}  & \small{0.99} & 	\small{WormsTwoClass} & \small{\textbf{0.64}} & \small{2.77} & \small{\textbf{0.11}}  & \small{18.75} \tabularnewline
\small{Meat} & \small{\textbf{2.10}} & \small{3.32} & \small{0.72} & \small{\textbf{0.71}} & 	\small{yoga} & \small{\textbf{1.23}} & \small{5.26} & \small{\textbf{0.65}}  & \small{2.92} \tabularnewline
\bottomrule
\end{tabular} }

\caption{\label{tab:Mean-absolute-deviation}Alignment errors of shapeDTW vs
dDTW. We use training data from each UCR dataset as the original time
series, and simulate alignment pairs by scaling and streching the
original time series (stretched by $30\%$). Then we run shapeDTW
and dDTW to align these synthesized alignment pairs, and evaluate
the alignment paths against the ground-truth by computing ``\textit{Mean Absolute Deviation}''
scores. The mean and standard deviation of the ``\textit{Mean Absolute Deviation}''
scores on each dataset is documented, with smaller means and stds
in bold font. shapeDTW achieves lower ``\textit{Mean Absolute Deviation}''
scores than dDTW on 56 datasets, showing its clear advantage for time
series alignment.}
\end{table*}

From the above simulation experiments, we observe dDTW (under the
step patterns ``symmetric1'' and ``asymmetric'') has the closest performance
as shapeDTW. Here we simulate aligned-pairs with on average $30\%$
stretches, run dDTW (under ``symmetric1'' step pattern) and shapeDTW
alignments, and report the ``\textit{Mean Absolute Deviation}'' scores
in Table \ref{tab:Mean-absolute-deviation}. shapeDTW has lower ``\textit{Mean Absolute Deivation}''
scores on 56 datasets, and the mean of ``\textit{Mean Absolute Deivation}''
on 84 datasets of shapeDTW and dDTW are 1.68/2.75 respectively, indicating
shapeDTW achieves much lower alignment errors. This shows a clear
superiority of shapeDTW to dDTW for sequence alignment.

The key difference between shapeDTW and DTW/dDTW/wDTW is that whether
neighborhood is taken into account when measuring similarities between
two points. We demonstrate that taking local neighborhood information
into account (shapeDTW) does benefit the alignment.

Notes: before running shapeDTW and DTW variants alignment, two sequences
in a simulated pair are z-normalized in advance; when computing ``\textit{Mean Absolute Deviation}'',
we choose the original time series as the reference sequence, i.e.,
divide the area between two alignment paths by the length of the original
time series.

\begin{figure*}[htbp!]
\begin{centering}
\includegraphics[width=1.0\textwidth]{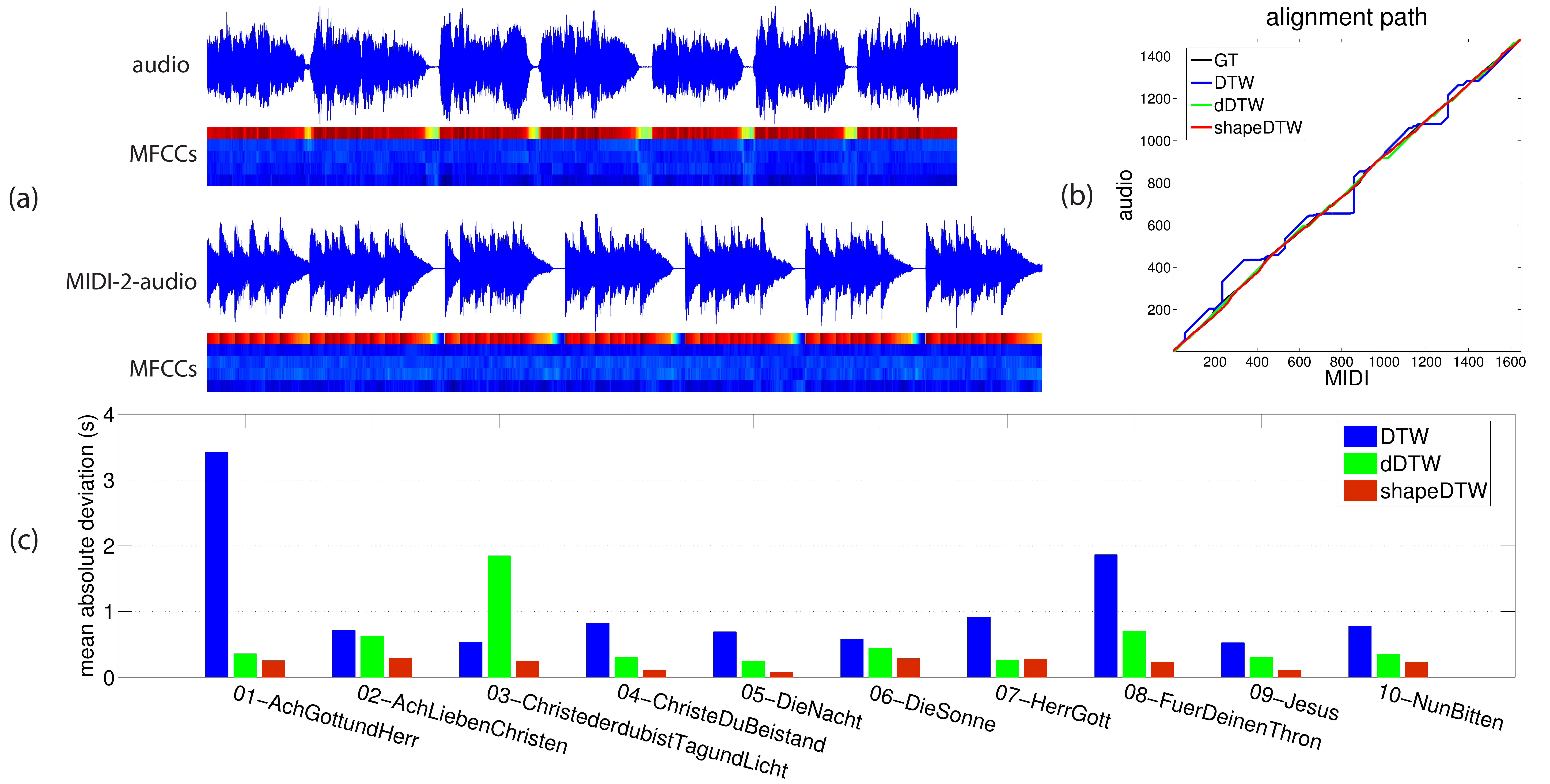} 
\par\end{centering}

\caption{\label{fig:fig-align-midi2audio} Align audio and midi-2-audio sequences.
(a) top: the audio waveform of the Chorale '05-DieNacht' and its 5D
MFCCs features; bottom: the converted audio waveform from the MIDI
score of the Chorale '05-DieNacht' and its corresponding 5D MFCCs
features; (b) alignment paths: align two MFCCs sequences by DTW, dDTW
and shapeDTW, and the plot shows their alignment paths, together with
the ground-truth alignment. As seen, the alignment paths of dDTW and
shapeDTW are closer to the ground-truth than that of DTW. (c) ``\textit{Mean Absolute Deviation}''
from the ground truth alignment: on 9 (10) out of 10 chorales, shapeDTW
achieves smaller alignment errors than dDTW (DTW), showing that shapeDTW
outperforms DTW/dDTW to align real sequence pairs as well.}
\end{figure*}

\subsubsection{\label{sub:MIDI-to-audio-alignment}MIDI-to-audio alignment}

We showed the superiority of shapeDTW to align synthesized alignment
pairs, and in this section, we further empirically demonstrate its
effectiveness to align audio signals, which have ground-truth alignments.

The Bach10 dataset \cite{duan2011soundprism} consists of audio recordings
of 10 pieces of Bach's Chorales, as well as their MIDI scores and
the ground-truth alignment between the audio and the MIDI score. MIDI
scores are symbolic representations of audio files, and by aligning
symbolic MIDI scores with audio recordings, we can do musical information
retrieval from MIDI input-data \cite{hu2003polyphonic}. Many previous
work used DTW to align MIDI to audio sequences \cite{hu2003polyphonic,duan2011soundprism,garreau2014metric},
and they typically converted MIDI data into audios as a first step,
and the problem boils down to audio-to-audio alignment, which is then
solved by DTW. We follow this convention to convert MIDI to audio
first, but run shapeDTW instead for alignments.

Each piece of music is approximately 30 seconds long, and in experiments,
we segment both the audio and the converted audio from MIDI data into
frames of 46ms length with a hopsize of 23ms, extract features from
each 46ms frame window, and in this way, the audio is represented
as a multivariate time series with the length equal to the number
of frames and dimension equal to the feature dimensions. There are
many potential choices of frame features, but how to select and combine
features in an optimal way to improve the alignment is beyond the
scope of this paper, we refer the interested readers to \cite{kirchhoff2011evaluation,garreau2014metric}.
Without loss of generality, we use Mel-frequency cepstral coefficients
(MFCCs) as features, due to its common usage and good performance
in speech recognition and musical information retrieval \cite{logan2000mel}.
In our experiments, we use the first 5 MFCCs coefficients.

After MIDI-to-audio conversion and MFCCs feature extraction, MIDI
files and audio recordings are represented as 5-dimensional multivariate
time series, with approximately length $L\approx1300$. A typical
audio signal, MIDI-converted audio signal, and their 5D MFCCs features
are shown in Fig. \ref{fig:fig-align-midi2audio}. We align 5D MFCCs
sequences by shapeDTW: although shapeDTW is designed for univariate
time series alignments, it naturally extends to multivariate cases:
first extract a subsequence from each temporal point, then encode
subsequences by shape descriptors, and in this way, the raw multivariate
time series is converted to a descriptor sequence. In the multivariate
time series case, each extracted subsequence is multi-dimensional,
having the same dimension as the raw time series, and to compute the
shape descriptor of a multi-dimensional subsequence, we compute shape
descriptors of each dimension independently, concatenate all shape descriptors, and use it as the shape representation of that
subsequence.

We compare alignments by shapeDTW against DTW/dDTW, and all of them
use the ``symmetric1'' step pattern. The length of subsequences in shapeDTW
is fixed to be 20 (we tried 5,10, 30 as well and achieved quite similar
results), and $\textit{Derivative}$ is used as the shape descriptor.
The alignment qualities in terms of ``\textit{Mean Absolute Deviation}''
on 10 Chorales are plotted in Fig. \ref{fig:fig-align-midi2audio}.
To be consistent with the convention in the audio community, we actually
report the mean-delayed-second between the alignment paths and the
ground-truth. The mean-delayed-second is computed as: dividing ``\textit{Mean Absolute Deviation}''
by the sampling rate of the audio signal. shapeDTW outperforms dDTW/DTW
on 9/10 MIDI-to-audio alignments. This shows taking local neighborhood
information into account does benefit the alignment.

\begin{figure*}[htbp!]
\begin{centering}
\includegraphics[width=1.0\textwidth]{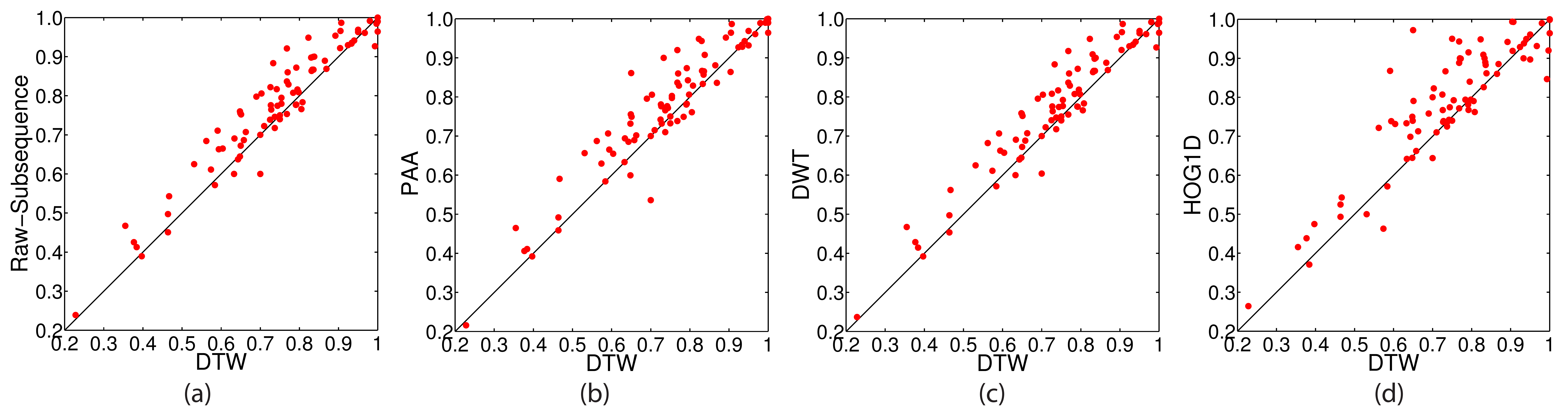} 
\par\end{centering}

\caption{\label{fig:fig-scatter} Classification accuracy comparisons between
NN-DTW and NN-shapeDTW on 84 UCR time series datasets. shapeDTW under
4 shape descriptors, $\textit{Raw-Subsequence}$, $\textit{PAA}$,
$\textit{DWT}$ and $\textit{HOG1D}$, outperforms DTW on 64/63/64/61
datasets respectively, and Wilcoxon signed rank test shows shapeDTW
under all descriptors performs significantly better than DTW. $\textit{Raw-Subsequence}$
($\textit{PAA}$ and $\textit{DWT}$ as well) outperforms DTW on more
datasets than $\textit{HOG1D}$ does, but $\textit{HOG1D}$ achieves
large accuracy improvements on more datasets, concretely, $\textit{HOG1D}$
boosts accuracies by more than $10\%$ on 18 datasets, compared with
on 12 datasets by $\textit{Raw-Subsequence}$.}
\end{figure*}

\subsection{Time series classification}

We compare NN-shapeDTW with NN-DTW on 84 UCR time series datasets
for classification. Since these datasets have standard partitions
of training and test data, we experiment with these given partitions
and report classification accuracies on the test data.

In the above section, we explore the influence of different steps patterns, but here both DTW and shapeDTW use the widely adopted 
step pattern ``symmetric1''(Fig. \ref{fig:Five-step-patterns.} (a)) under no temporal window constraints to align sequences.

\noindent $\textbf{NN-DTW}$: each test time series is compared against
the training set, and the label of the training time series with the
minimal DTW distance to that test time series determines the predicted
label. All training and testing time series are z-normalized in advance.

\noindent $\textbf{shapeDTW}$: we test all 5 shape descriptors. We
z-normalize time series in advance, sample subsequences from the time
series, and compute 3 magnitude-aware shape descriptors, $\textit{Raw-Subsequence}$,
$\textit{PAA}$ and $\textit{DWT}$, and 2 y-shift invariant shape
descriptors, $\textit{Slope}$ and $\textit{HOG1D}$. Parameter setting
for 5 shape descriptors: (1) The length of subsequences to be sampled
around temporal points is fixed to 30, as a result $\textit{Raw-Subsequence}$
descriptor is a 30D vector; (2) $\textit{PAA}$ and $\textit{Slope}$
uses 5 equal-lengthed intervals, therefore they have the dimensionality
5; (3) As mentioned, $\textit{HOG1D}$ uses 8 bins and 2 non-overlapping
intervals, and the scale factor $\sigma$ is fixed to be 0.1. At last HOG1D is a 16D vector representation.

\noindent $\textbf{NN-shapeDTW}$: first transform each training/testing
time series to a shape descriptor sequence, and in this way, original
univariate time series are converted into multivariate descriptor
time series. Then apply NN-DTW on the multivariate time series to
predict labels.

$\textbf{NN-shapeDTW vs. NN-DTW}$: we compare NN-shapeDTW, under
4 shape descriptors $\textit{Raw-Subsequence}$, $\textit{PAA}$,
$\textit{DWT}$ and $\textit{HOG1D}$, with NN-DTW, and plot their
classification accuracies on 84 datasets in Fig.\ref{fig:fig-scatter}.
shapeDTW outperforms (including ties) DTW on 64/63/64/61 ($\textit{Raw-Subsequence}$/$\textit{PAA}$/$\textit{DWT}$/$\textit{HOG1D}$)
datasets, and by running the Wilcoxon signed rank test between performances
of NN-shapeDTW and NN-DTW, we obtain p-values $5.5\cdot10^{-8}$/$5.1\cdot10^{-7}$/$4.8\cdot10^{-8}$/$1.7\cdot10^{-6}$,
showing that shapeDTW under all 4 descriptors performs significantly
better than DTW. Compared with DTW, shapeDTW has a preceding shape
descriptor extraction process, and approximately takes time $\mathcal{O}(l\cdot L)$,
where $l$ and $L$ is the length of subsequence and time series respectively.
Since generally $l\ll L$, the total time complexity of shapeDTW is
$\mathcal{O}(L^{2})$, which is the same as DTW. By trading off a
slight amount of time and space, shapeDTW brings large accuracy gains.

Since $\textit{PAA}$ and $\textit{DWT}$ are approximations of \textit{Raw-Subsequence},
and they have similar performances as \textit{Raw-Subsequence}
under the nearest classifier, we choose \textit{Raw-Subsequence}
as a representative for following analysis. Shape descriptor \textit{Raw-Subsequence}
loses on 20 datasets, on 18 of which it has minor losses ($<4\%$),
and on the other 2 datasets, ``Computers'' and ``Synthetic-control'',
it loses by $10\%$ and $6.6\%$. Time series instances from these
2 datasets either have high-frequency spikes or have many abrupt direction
changes, making them resemble noisy signals very much. Possibly, comparing
the similarity of two points using their noisy neighborhoods is not
as good as using their single coordinate values (DTW), since temporal
neighborhood may accumulate and magnify noise.

$\textit{HOG1D}$ loses on 23 datasets, on 18 of which it has minor
losses ($<5\%$), and on the other 5 datasets, ``CBF'', ``Computers'',
``ItalyPowerDemand'', ``Synthetic-control'' and ``Wine'', it loses by $7.7\%$,
$5.6\%$, $5.3\%$, $14\%$ and $11\%$. By visually inspecting, time
series from ``Computers'', ``CBF'' and ``Synthetic-control'' are spiky and
bumpy, making them highly non-smooth. This makes the first-order-derivative
based descriptor $\text{\textit{HOG1D}}$ inappropriate to represent
local structures. Time series instances from 'ItalyPowerDemand' have
length 24, while we sample subsequences of length 30 from each point,
this makes $\textit{HOG1D}$ descriptors from different local points
almost the same, such that $\textit{HOG1D}$ becomes not discriminative
of local structures. This makes shapeDTW inferior to DTW. Although
$\textit{HOG1D}$ loses on more datasets than \textit{Raw-Subsequence},
$\textit{HOG1D}$ boosts accuracies by more than $10\%$ on 18 datasets,
compared with on 12 datasets by \textit{Raw-Subsequence}. On datasets
``OSUleaf'' and ``BirdChicken'', the accuracy gain is as high as $27\%$
and $20\%$. By checking these two datasets closely, we find different
classes have membership-discriminative local patterns (a.k.a shapelets
\cite{ye2009time}), however, these patterns differ only slightly
among classes. \textit{Raw-Subsequence} shape descriptor can not
capture these minor differences well, while $\textit{HOG1D}$ is more
sensitive to shape variations since it calculates derivatives.

Both \textit{Raw-Subsequence} and $\textit{HOG1D}$ bring significant
accuracy gains, however, they boost accuracies to different extents
on the same dataset. This indicates the importance of designing domain-specific
shape descriptors. Nevertheless, we show that even by using simple
and dataset-independent shape descriptors, we still obtain significant
improvements over DTW. Classification error rates of DTW, \textit{Raw-Subsequence}
and $\textit{HOG1D}$ on 84 datasets are documented in Table.\ref{errorrates}.

\begin{figure}[h]
\begin{centering}
\includegraphics[width=0.49\textwidth]{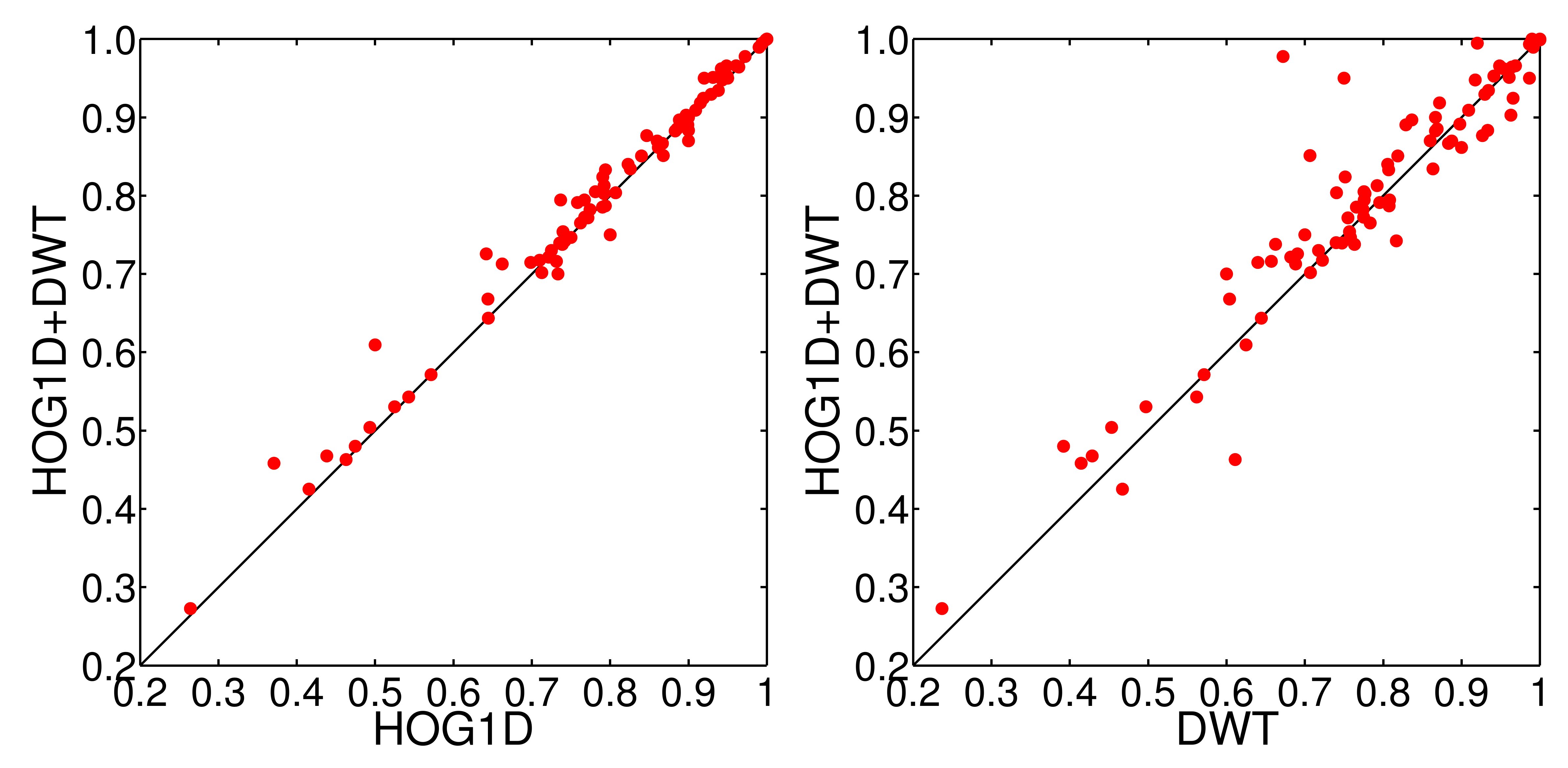} 
\par\end{centering}

\caption{\label{fig:fig-complementary} Performance comparisons between the
fused descriptor $\textit{HOG1D+DWT}$ and individual ones $\textit{HOG1D}$/$\textit{DWT}$.
$\textit{HOG1D+DWT}$ outperforms $\textit{HOG1D}$/$\textit{DWT}$
on 66/51 (out of 84) datasets, and statistical hypothesis tests show
the improvements are significant.}
\end{figure}

$\textbf{Superiority of Compound shape descriptors}$: as mentioned
in Sec.\ref{sec:shape-descriptors}, a compound shape descriptor obtained
by fusing two complementary descriptors may inherit benefits from
both descriptors, and becomes even more discriminative of subsequences.
As an example, we concatenate a y-shift invariance descriptor \textit{HOG1D}
and a magnitude-aware descriptor \textit{DWT} using equal weights,
resulting in a compound descriptor $\textit{HOG1D}+\textit{DWT}=(\textit{HOG1D},\:\textit{DWT})$.
Then we evaluate classification performances of 3 descriptors under
the nearest neighbor classifier, and plot the comparisons in Fig.\ref{fig:fig-complementary}.
$\textit{HOG1D+DWT}$ outperforms (including ties) $\textit{HOG1D}$ / $\textit{DWT}$
on 66/51 (out of 84) datasets, and by running the Wilcoxon signed
rank hypothesis test between performances of $\textit{HOG1D+DWT}$
and $\textit{HOG1D}$ ($\textit{DWT}$), we get p-values $5.5\cdot10^{-5}$/$0.0034$,
showing the compound descriptor outperforms individual descriptors
significantly under the confidence level $5\%$. We can generate compound
descriptors by weighted concatenation, with weights tuned by cross-validation
on training data, but this is beyond the scope of this paper.

\begin{figure}[h]
\begin{centering}
\includegraphics[width=0.50\textwidth]{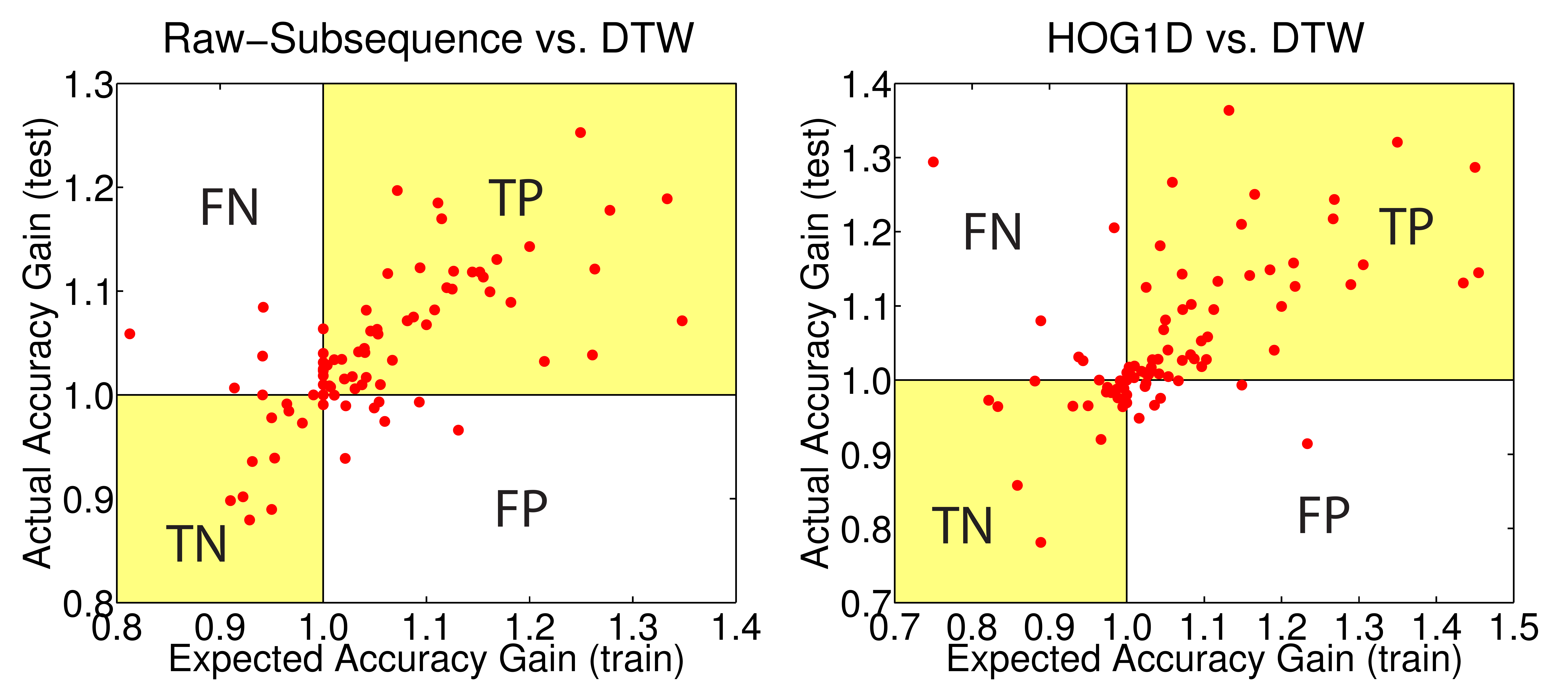} 
\par\end{centering}

\caption{\label{fig:fig-texas} Texas sharpshoot plot between $\textit{Raw-Subsequence}$/$\textit{HOG1D}$
and DTW on 84 datasets. TP: true positive (our algorithm was expected
from the training data to outperform NNDTW, and it actually did on
the test data). TN: true negatives, FP: false positives, FN: false
negatives. There are $87\%$/$86\%$ points ($\textit{Raw-Subsequence}$/$\textit{HOG1D}$
vs. DTW) falling in the TP and TN regions, which indicates we can
confidently predict that our algorithm will be superior/inferior to
NNDTW. }
\end{figure}

$\textbf{Texas Sharpshooter plot:}$ although NN-shapeDTW performs better than
NN-DTW, knowing this is not useful unless we can tell in advance on
which problems it will be more accurate, as stated in \cite{batista2011complexity}.
Here we use the Texas sharpshooter plot \cite{batista2011complexity}
to show when NN-shapeDTW has superior performance on the test set
as predicted from performance on the training set, compared with NN-DTW.
We run leave-one-out cross validation on training data to measure
the accuracies of NN-shapeDTW and NN-DTW, and we calculate the expected
gain: accuracy(NN-shapeDTW)/accuracy(NN-DTW). We then measure the
actual accuracy gain using the test data. The Texas Sharpshooter plots
between \textit{Raw-Subsequence}/\textit{HOG1D} and DTW on 84
datasets are shown in Fig.\ref{fig:fig-texas}. $87\%$/$86\%$ points
(\textit{Raw-Subsequence}/\textit{HOG1D}) fall in the TP and
TN regions, which means we can confidently predict that our algorithm
will be superior/inferior to NNDTW. There are respectively 7/7 points
falling inside the FP region for descriptors \textit{Raw-Subsequence}/\textit{HOG1D},
but they just represent minor losses, i.e., actual accuracy gains
lie within $[0.9\:1.0]$.

\vspace{-12pt}

\subsection{Sensitivity to the size of neighborhood}

In the above experiments, we showed that shapeDTW outperforms DTW
both qualitatively and quantitatively. But we are still left with
one free-parameter: the size of neighborhood, i.e., the length of
the subsequence to be sampled from each point. Let $t_{i}$ be some
temporal point on the time series $\mathcal{T}\in\mathcal{R}^{L}$,
and $s_{i}$ be the subsequence sampled at $t_{i}$. When $|s_{i}|=1$,
shapeDTW (under the $\textit{Raw-Subsequence}$ shape descriptor)
degenerates to DTW; when $|s_{i}|=L$, subsequences sampled at different
points become almost identical, make points un-identifiable by their
shape descriptors. This shows the importance to set an appropriate
subsequence length. However, without dataset-specific domain knowledge,
it is hard to determine the length intelligently. Here instead, we
explore the sensitivity of the classification accuracies to different
subsequence lengths. We conduct experiments on 42 old UCR datasets.

\begin{figure*}[htbp!]
\begin{centering}
\includegraphics[width=1.0\textwidth]{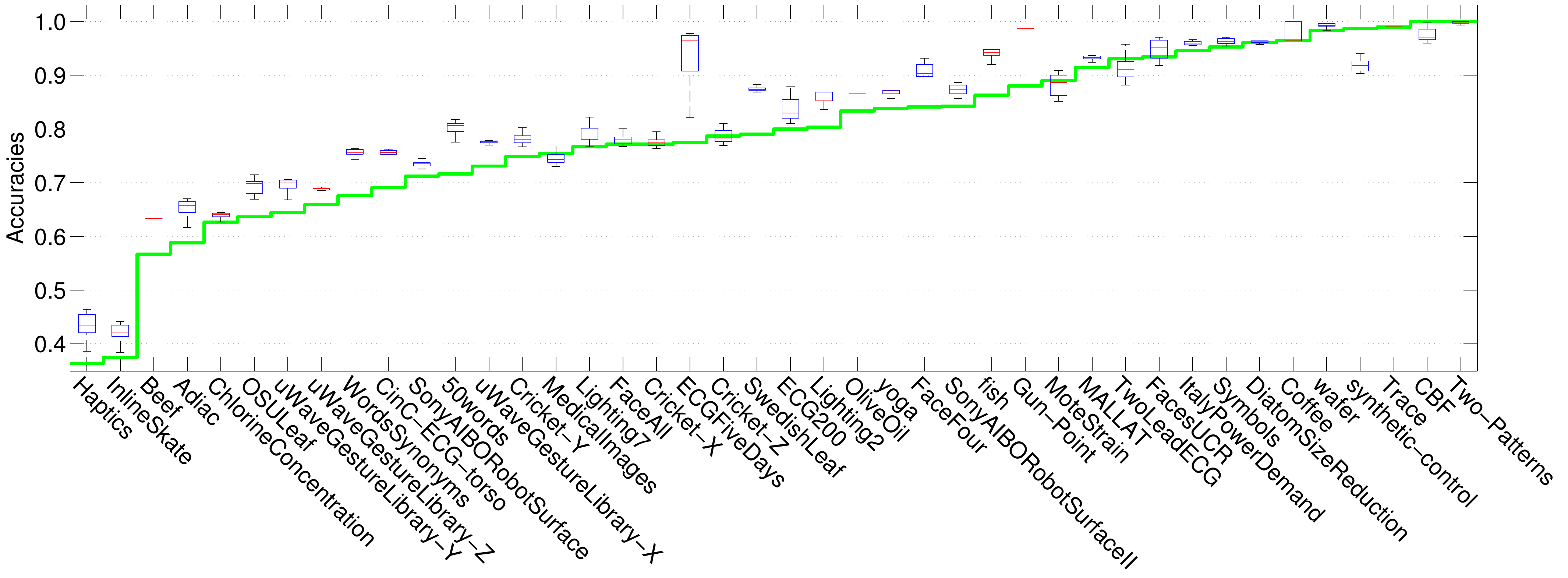} 
\par\end{centering}

\caption{\label{fig:fig-insensitivity2seqlen} Performances of shapeDTW are
insensitive to the neighborhood size. The green stairstep curve shows
dataset-wise test accuracies of NN-DTW, and the box plot shows performances
of NN-shapeDTW under the shape descriptor $\textit{Raw-Subsequence}$.
On each dataset, we plot a blue box with two tails: the lower and
upper edge of each blue box represent $25th$ and $75th$ percentiles
of 20 test accuracies (obtained under different neighborhood sizes,
i.e., 5, 10, 15, ..., 100) on that dataset, with the red line inside
the box marking the median accuracy and two tails indicating the best
and worst test accuracies. On 36 out of 42 datasets, the median accuracies
of NN-shapeDTW are larger than accuracies obtained by NN-DTW, and
on 33 datasets, even the worst performances by NN-shapeDTW are better
than NN-DTW. All these statistics show shapeDTW works well under wide
ranges of neighborhood sizes.}
\end{figure*}

We use $\textit{Raw-Subsequence}$ as the shape descriptor, and NN-shapeDTW
as the classifier. We let the length of subsequences to vary from
5 to 100, with stride 5, i.e., we repeat classification experiments
on each dataset for 20 times, and each time set the length of subsequences
to be $5\times i$, where $i$ is the index of experiments ($1\le i\le20,\, i\in\mathcal{Z}$).
The test accuracies under 20 experiments are shown by a box plot (
Fig.\ref{fig:fig-insensitivity2seqlen}). On 33 out of 42 datasets,
even the worst performances of NN-shapeDTW are better than DTW, indicating
shapeDTW performs well under wide ranges of neighborhood sizes.

\begin{table*}[!htbp]
\global\long\def\tabincell#1#2{\begin{tabular}{@{}@{}}
 #2\end{tabular}}
 \centering %
 \resizebox{2.0\columnwidth}{!} {
\begin{tabular}{lccclccc}
\toprule 
\multicolumn{8}{c}{\textbf{\normalsize{classification
error rates on 84 UCR datasets}}}\tabularnewline
\midrule 
{\normalsize{}{}{}{datasets} }  & {\normalsize{}{}{}{DTW} }  & {\normalsize{}{}{}{Raw-Subsequence} }  & {\normalsize{}{}{}{HOG1D} }  & {\normalsize{}{}{}{datasets} }  & {\normalsize{}{}{}{DTW} }  & {\normalsize{}{}{}{Raw-Subsequence} }  & {\normalsize{}{}{}{HOG1D} }\tabularnewline
\midrule 
\normalsize{50words} & \normalsize{0.310} & \underline{\normalsize{\textbf{0.202}}} &\normalsize{\textbf{0.242}} & 	 \normalsize{MedicalImages} &  \normalsize{0.263} & \normalsize{\textbf{0.254}} &\normalsize{0.264} \tabularnewline
\normalsize{Adiac} & \normalsize{0.396} & \normalsize{\textbf{0.335}} &\underline{\normalsize{\textbf{0.269}}} & 	 \normalsize{MiddlePhalanxOutlineAgeGroup} & \normalsize{\textbf{0.250}} & \normalsize{0.260} &\normalsize{0.260} \tabularnewline
\normalsize{ArrowHead} & \normalsize{0.297} & \underline{\normalsize{\textbf{0.194}}} &\underline{\normalsize{\textbf{0.177}}} & 	 \normalsize{MiddlePhalanxOutlineCorrect} &  \normalsize{0.352} & \underline{\normalsize{\textbf{0.240}}} &\underline{\normalsize{\textbf{0.250}}}\tabularnewline
\normalsize{Beef} & \normalsize{0.367} & \normalsize{0.400} &\underline{\normalsize{\textbf{0.267}}} & 	 \normalsize{MiddlePhalanxTW} & \normalsize{\textbf{0.416}} & \normalsize{0.429} &\normalsize{0.429} \tabularnewline
\normalsize{BeetleFly} & \normalsize{0.300} & \normalsize{\textbf{0.300}} &\underline{\normalsize{\textbf{0.200}}} & 	 \normalsize{MoteStrain} &  \normalsize{0.165} & \normalsize{\textbf{0.101}} &\normalsize{\textbf{0.110}} \tabularnewline
\normalsize{BirdChicken} & \normalsize{0.250} & \normalsize{\textbf{0.250}} &\underline{\normalsize{\textbf{0.050}}} & 	 \normalsize{NonInvasiveFatalECG-Thorax1} & \normalsize{\textbf{0.209}} & \normalsize{0.223} &\normalsize{0.219} \tabularnewline
\normalsize{Car} & \normalsize{0.267} & \underline{\normalsize{\textbf{0.117}}} &\underline{\normalsize{\textbf{0.133}}} & 	 \normalsize{NonInvasiveFatalECG-Thorax2} &  \normalsize{0.135} & \normalsize{\textbf{0.110}} &\normalsize{0.140} \tabularnewline
\normalsize{CBF} & \normalsize{\textbf{0.003}} & \normalsize{0.016} &\normalsize{0.080} & 	 \normalsize{OliveOil} &  \normalsize{0.167} & \normalsize{\textbf{0.133}} &\normalsize{\textbf{0.100}} \tabularnewline
\normalsize{ChlorineConcentration} & \normalsize{\textbf{0.352}} & \normalsize{0.355} &\normalsize{0.355} & 	 \normalsize{OSULeaf} &  \normalsize{0.409} & \underline{\normalsize{\textbf{0.289}}} &\underline{\normalsize{\textbf{0.132}}}\tabularnewline
\normalsize{CinC-ECG-torso} & \normalsize{0.349} & \underline{\normalsize{\textbf{0.248}}} &\underline{\normalsize{\textbf{0.209}}} & 	 \normalsize{PhalangesOutlinesCorrect} &  \normalsize{0.272} & \normalsize{\textbf{0.235}} &\normalsize{\textbf{0.261}} \tabularnewline
\normalsize{Coffee} & \normalsize{\textbf{0.000}} & \normalsize{0.036} &\normalsize{0.036} & 	 \normalsize{Phoneme} &  \normalsize{0.772} & \normalsize{\textbf{0.761}} &\normalsize{\textbf{0.736}} \tabularnewline
\normalsize{Computers} & \normalsize{\textbf{0.300}} & \normalsize{0.400} &\normalsize{0.356} & 	 \normalsize{Plane} &  \normalsize{0.000} & \normalsize{\textbf{0.000}} &\normalsize{\textbf{0.000}} \tabularnewline
\normalsize{Cricket-X} & \normalsize{0.246} & \normalsize{\textbf{0.221}} &\normalsize{\textbf{0.208}} & 	 \normalsize{ProximalPhalanxOutlineAgeGroup} & \normalsize{\textbf{0.195}} & \normalsize{0.234} &\normalsize{0.210} \tabularnewline
\normalsize{Cricket-Y} & \normalsize{0.256} & \normalsize{\textbf{0.226}} &\normalsize{\textbf{0.226}} & 	 \normalsize{ProximalPhalanxOutlineCorrect} &  \normalsize{0.216} & \normalsize{\textbf{0.192}} &\normalsize{\textbf{0.206}} \tabularnewline
\normalsize{Cricket-Z} & \normalsize{0.246} & \normalsize{\textbf{0.205}} &\normalsize{\textbf{0.208}} & 	 \normalsize{ProximalPhalanxTW} & \normalsize{\textbf{0.263}} & \normalsize{0.282} &\normalsize{0.275} \tabularnewline
\normalsize{DiatomSizeReduction} & \normalsize{\textbf{0.033}} & \normalsize{0.039} &\normalsize{0.069} & 	 \normalsize{RefrigerationDevices} &  \normalsize{0.536} & \normalsize{0.549} &\normalsize{\textbf{0.507}} \tabularnewline
\normalsize{DistalPhalanxOutlineAgeGroup} & \normalsize{\textbf{0.208}} & \normalsize{0.223} &\normalsize{0.233} & 	 \normalsize{ScreenType} &  \normalsize{0.603} & \normalsize{0.611} &\normalsize{\textbf{0.525}} \tabularnewline
\normalsize{DistalPhalanxOutlineCorrect} & \normalsize{0.232} & \normalsize{0.247} &\normalsize{\textbf{0.228}} & 	 \normalsize{ShapeletSim} &  \normalsize{0.350} & \normalsize{\textbf{0.328}} &\underline{\normalsize{\textbf{0.028}}}\tabularnewline
\normalsize{DistalPhalanxTW} & \normalsize{0.290} & \normalsize{\textbf{0.277}} &\normalsize{\textbf{0.290}} & 	 \normalsize{ShapesAll} &  \normalsize{0.232} & \normalsize{\textbf{0.163}} &\underline{\normalsize{\textbf{0.112}}}\tabularnewline
\normalsize{Earthquakes} & \normalsize{0.258} & \normalsize{\textbf{0.183}} &\normalsize{\textbf{0.258}} & 	 \normalsize{SmallKitchenAppliances} &  \normalsize{0.357} & \normalsize{0.363} &\normalsize{\textbf{0.301}} \tabularnewline
\normalsize{ECG200} & \normalsize{0.230} & \normalsize{\textbf{0.140}} &\underline{\normalsize{\textbf{0.100}}} & 	 \normalsize{SonyAIBORobotSurface} &  \normalsize{0.275} & \normalsize{\textbf{0.261}} &\normalsize{\textbf{0.193}} \tabularnewline
\normalsize{ECG5000} & \normalsize{0.076} & \normalsize{\textbf{0.070}} &\normalsize{\textbf{0.071}} & 	 \normalsize{SonyAIBORobotSurfaceII} &  \normalsize{0.169} & \normalsize{\textbf{0.136}} &\normalsize{0.174} \tabularnewline
\normalsize{ECGFiveDays} & \normalsize{0.232} & \underline{\normalsize{\textbf{0.079}}} &\underline{\normalsize{\textbf{0.057}}} & 	 \normalsize{Strawberry} &  \normalsize{0.060} & \normalsize{\textbf{0.059}} &\normalsize{\textbf{0.051}} \tabularnewline
\normalsize{FaceAll} & \normalsize{\textbf{0.192}} & \normalsize{0.217} &\normalsize{0.238} & 	 \normalsize{SwedishLeaf} &  \normalsize{0.208} & \normalsize{\textbf{0.128}} &\underline{\normalsize{\textbf{0.085}}}\tabularnewline
\normalsize{FaceFour} & \normalsize{0.170} & \normalsize{\textbf{0.102}} &\normalsize{\textbf{0.091}} & 	 \normalsize{Symbols} &  \normalsize{0.050} & \normalsize{\textbf{0.031}} &\normalsize{\textbf{0.039}} \tabularnewline
\normalsize{FacesUCR} & \normalsize{0.095} & \normalsize{\textbf{0.034}} &\normalsize{\textbf{0.081}} & 	 \normalsize{synthetic-control} & \normalsize{\textbf{0.007}} & \normalsize{0.073} &\normalsize{0.153} \tabularnewline
\normalsize{FISH} & \normalsize{0.177} & \underline{\normalsize{\textbf{0.051}}} &\underline{\normalsize{\textbf{0.051}}} & 	 \normalsize{ToeSegmentation1} &  \normalsize{0.228} & \normalsize{\textbf{0.171}} &\underline{\normalsize{\textbf{0.101}}}\tabularnewline
\normalsize{FordA} & \normalsize{0.438} & \underline{\normalsize{\textbf{0.316}}} &\underline{\normalsize{\textbf{0.279}}} & 	 \normalsize{ToeSegmentation2} &  \normalsize{0.162} & \normalsize{\textbf{0.100}} &\normalsize{\textbf{0.138}} \tabularnewline
\normalsize{FordB} & \normalsize{0.406} & \normalsize{\textbf{0.337}} &\underline{\normalsize{\textbf{0.261}}} & 	 \normalsize{Trace} &  \normalsize{0.000} & \normalsize{0.010} &\normalsize{\textbf{0.000}} \tabularnewline
\normalsize{Gun-Point} & \normalsize{0.093} & \normalsize{\textbf{0.013}} &\normalsize{\textbf{0.007}} & 	 \normalsize{TwoLeadECG} &  \normalsize{0.096} & \normalsize{\textbf{0.078}} &\normalsize{\textbf{0.006}} \tabularnewline
\normalsize{Ham} & \normalsize{0.533} & \normalsize{\textbf{0.457}} &\normalsize{\textbf{0.457}} & 	 \normalsize{Two-Patterns} &  \normalsize{0.000} & \normalsize{\textbf{0.000}} &\normalsize{0.001} \tabularnewline
\normalsize{HandOutlines} & \normalsize{0.202} & \normalsize{\textbf{0.191}} &\normalsize{0.206} & 	 \normalsize{UWaveGestureLibraryAll} &  \normalsize{0.108} & \normalsize{\textbf{0.046}} &\normalsize{\textbf{0.058}} \tabularnewline
\normalsize{Haptics} & \normalsize{0.623} & \normalsize{\textbf{0.575}} &\normalsize{\textbf{0.562}} & 	 \normalsize{uWaveGestureLibrary-X} &  \normalsize{0.273} & \normalsize{\textbf{0.224}} &\normalsize{\textbf{0.263}} \tabularnewline
\normalsize{Herring} & \normalsize{0.469} & \normalsize{\textbf{0.375}} &\normalsize{0.500} & 	 \normalsize{uWaveGestureLibrary-Y} &  \normalsize{0.366} & \normalsize{\textbf{0.309}} &\normalsize{\textbf{0.358}} \tabularnewline
\normalsize{InlineSkate} & \normalsize{0.616} & \normalsize{\textbf{0.587}} &\normalsize{0.629} & 	 \normalsize{uWaveGestureLibrary-Z} &  \normalsize{0.342} & \normalsize{\textbf{0.314}} &\normalsize{\textbf{0.338}} \tabularnewline
\normalsize{InsectWingbeatSound} & \normalsize{0.645} & \underline{\normalsize{\textbf{0.533}}} &\normalsize{\textbf{0.584}} & 	 \normalsize{wafer} &  \normalsize{0.020} & \normalsize{\textbf{0.008}} &\normalsize{\textbf{0.010}} \tabularnewline
\normalsize{ItalyPowerDemand} & \normalsize{0.050} & \normalsize{\textbf{0.037}} &\normalsize{0.103} & 	 \normalsize{Wine} &  \normalsize{0.426} & \normalsize{\textbf{0.389}} &\normalsize{0.537} \tabularnewline
\normalsize{LargeKitchenAppliances} & \normalsize{0.205} & \normalsize{\textbf{0.184}} &\normalsize{\textbf{0.160}} & 	 \normalsize{WordsSynonyms} &  \normalsize{0.351} & \underline{\normalsize{\textbf{0.245}}} &\normalsize{\textbf{0.260}} \tabularnewline
\normalsize{Lighting2} & \normalsize{0.131} & \normalsize{0.131} &\normalsize{\textbf{0.115}} & 	 \normalsize{WordSynonyms} &  \normalsize{0.351} & \underline{\normalsize{\textbf{0.245}}} &\normalsize{\textbf{0.260}} \tabularnewline
\normalsize{Lighting7} & \normalsize{0.274} & \normalsize{\textbf{0.178}} &\normalsize{\textbf{0.233}} & 	 \normalsize{Worms} &  \normalsize{0.536} & \normalsize{\textbf{0.503}} &\normalsize{\textbf{0.475}} \tabularnewline
\normalsize{MALLAT} & \normalsize{0.066} & \normalsize{\textbf{0.064}} &\normalsize{\textbf{0.062}} & 	 \normalsize{WormsTwoClass} &  \normalsize{0.337} & \normalsize{\textbf{0.293}} &\normalsize{\textbf{0.287}} \tabularnewline
\normalsize{Meat} & \normalsize{0.067} & \normalsize{\textbf{0.067}} &\normalsize{0.100} & 	 \normalsize{yoga} &  \normalsize{0.164} & \normalsize{\textbf{0.133}} &\normalsize{\textbf{0.117}} \tabularnewline
\bottomrule
\end{tabular} }

\caption{\label{errorrates}Error rates of NN-DTW and NN-shapeDTW (under descriptors
$\textit{Raw-Subsequence}$ and $\textit{HOG1D}$) on 84 UCR datasets.
The error rates on datasets where NN-shapeDTW outperforms NN-DTW are
highlighted in bold font. Underscored datasets are those on which
shapeDTW has improved the accuracies by more than $10\%$.}
\end{table*}

\section{\label{sec:Conclusion}Conclusion}

We have proposed an new temporal sequence alignment algorithm, shapeDTW,
which achieves quantitatively better alignments than DTW and its variants.
shapeDTW is a quite generic framework as well, and uses can design
their own local subsequence descriptor and fit it into shapeDTW. We
experimentally showed that shapeDTW under the nearest neighbor classifier
obtains significantly improved classification accuracies than NN-DTW.
Therefore, NN-shapeDTW sets a new accuracy baseline for further comparison.




%

%

\ifCLASSOPTIONcompsoc
  \section*{Acknowledgments}
\else
  \section*{Acknowledgment}
\fi

This work was supported by the National Science Foundation (grant number CCF-1317433), the
Office of Naval Research (N00014-13-1-0563) and the Army Research Office (W911NF-11-1-0046 and W911NF-12-1-0433). The
authors affirm that the views expressed herein are solely their own, and do not represent the views of the United States
government or any agency thereof.

\ifCLASSOPTIONcaptionsoff
  \newpage
\fi



 \bibliographystyle{plain}
\bibliography{shapeDTW}

%
%
%
%
%

%

\begin{IEEEbiography}[{{\includegraphics[clip,width=1in,height=1.25in,keepaspectratio]{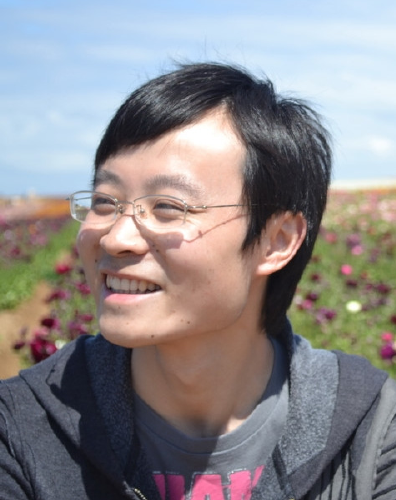}}}]
{Jiaping Zhao} received his bachelor and master degree from Wuhan
University in 2008 and 2010 respectively. Currently he is a Ph.D student
at iLab, University of Southern California, working under the supervision
of Laurent Itti. His research interests include computer vision, data
mining, visual attention and probabilistic graphical model. 
\end{IEEEbiography}

\begin{IEEEbiography}
[{\includegraphics[width=1in,height=1.25in,clip,keepaspectratio]{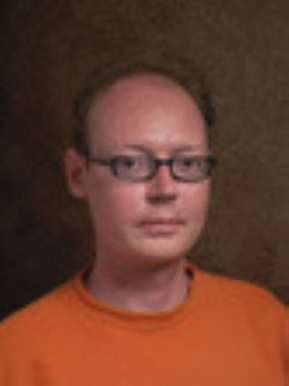}}]
{Laurent Itti} received the MS degree in image
processing from the Ecole Nationale Superiere
des Te´lecommunications in Paris in 1994 and
the PhD degree in computation and neural
systems from the California Institute of Technology
in 2000. Now he is an professor of
computer science, psychology, and neurosciences
at the University of Southern California.
His research interests include computational neuroscience, 
neural networks, visual attention, brain modelling.
He is a member of the IEEE.. 
\end{IEEEbiography}




\end{document}